\documentclass[11pt]{article}

\newif\ifsubmit
\submittrue

\usepackage[preprint]{acl} 

\usepackage{times}
\usepackage{latexsym}

\usepackage[T1]{fontenc}

\usepackage[utf8]{inputenc}

\usepackage{microtype}

\usepackage{inconsolata}

\usepackage{graphicx}

\usepackage[utf8]{inputenc} 
\usepackage[T1]{fontenc}    
\usepackage{url}            
\usepackage{booktabs}       
\usepackage{amsfonts}       
\usepackage{nicefrac}       
\usepackage{microtype}      
\usepackage{xcolor}         
\usepackage{graphicx}
\usepackage{array} 

\usepackage{tabularx}  
\usepackage{ragged2e}  
\usepackage{float} 
\usepackage[inline]{enumitem} 
\usepackage{cprotect} 
\usepackage{subcaption} 
\usepackage{siunitx}
\usepackage{minted} 
\usepackage{placeins} 
\usepackage[ruled,vlined]{algorithm2e}
\usepackage{amsmath}
\usepackage{listings} 

\usepackage{wrapfig} 

\usepackage[capitalize]{cleveref} 
\crefname{figure}{Fig.}{Figs.}
\crefname{table}{Table}{Tables}
\crefname{appendix}{App.}{Apps.}
\crefname{section}{\S}{\S\S}
\crefformat{section}{\S#2#1#3} 
\crefname{equation}{Eq.}{Eqs.}
\crefname{algorithm}{Alg.}{Algs.}
\crefname{algocf}{Alg.}{Algs.}
\crefname{defin}{Def.}{Defs.}
\crefname{theorem}{Thm.}{Thms.}
\crefname{lemma}{Lemma}{Lemmas}

\usepackage[toc,title,page]{appendix}
\usepackage{caption}   
\usepackage{mathtools} 

\ifsubmit
\usepackage[textsize=tiny,textwidth=1.3in,disable]{todonotes}
\else
\usepackage[textsize=scriptsize,textwidth=0.9in]{todonotes}
\fi

\usepackage{xurl} 
\usepackage{multirow} 

\newcommand{\andreasfinished}[2][]{} 

\newcommand{\maximilianfinished}[2][]{} 

\definecolor{codegreen}{rgb}{0,0.6,0}
\definecolor{codegray}{rgb}{0.5,0.5,0.5}
\definecolor{codepurple}{rgb}{0.58,0,0.82}
\definecolor{backcolour}{rgb}{0.95,0.95,0.92}

\lstdefinestyle{mystyle}{
    backgroundcolor=\color{backcolour},   
    commentstyle=\color{codegreen},
    keywordstyle=\color{magenta},
    numberstyle=\tiny\color{codegray},
    stringstyle=\color{codepurple},
    basicstyle=\ttfamily\footnotesize,
    breakatwhitespace=false,         
    breaklines=true,                 
    captionpos=b,                    
    keepspaces=true,                 
    numbers=left,                    
    numbersep=5pt,                  
    showspaces=false,                
    showstringspaces=false,
    showtabs=false,                  
    tabsize=2
}

\newcommand{\ood}{OOD\xspace}
\newcommand{\indist}{ID\xspace}
\newcommand{\genGapDef}{\ensuremath{\text{\indist} - \text{\ood}}}

\newcommand{\verl}{\texttt{VeRL}\xspace}

\newcommand{\noCurr}{\textit{Standard}\xspace}
\newcommand{\singleDiffInc}{\textit{SingleDiffInc}\xspace}
\newcommand{\singleDiffDec}{\textit{SingleDiffDec}\xspace}
\newcommand{\upToDiff}{\textit{UpToDiff}\xspace}
\newcommand{\downToDiff}{\textit{DownToDiff}\xspace}

\newcommand{\lindepth}{\textsc{LinearDepth}\xspace}
\newcommand{\partwhole}{\textsc{PartWhole}\xspace}

\newcommand{\mathgap}{\textsc{MathGAP}\xspace}
\newcommand{\kk}{\textsc{KK}\xspace}

\newcommand{\modelLOneF}{\textsc{Llama3.2-1B}\xspace}
\newcommand{\modelLThreeF}{\textsc{Llama3.2-3B}\xspace}
\newcommand{\modelQwenZeroSixF}{\textsc{Qwen3-0.6B}\xspace}
\newcommand{\modelQwenOneSevenF}{\textsc{Qwen3-1.7B}\xspace}
\newcommand{\modelQwenFourF}{\textsc{Qwen3-4B}\xspace}
\newcommand{\modelGemmaNineF}{\textsc{Gemma2-9B-Instruct}\xspace}
\newcommand{\QwenThreeSeries}{\textsc{Qwen3}\xspace}

\newcommand{\modelLOne}{\textsc{L1B}\xspace}
\newcommand{\modelLThree}{\textsc{L3B}\xspace}
\newcommand{\modelQwenZeroSix}{\textsc{Q0.6B}\xspace}
\newcommand{\modelQwenOneSeven}{\textsc{Q1.7B}\xspace}
\newcommand{\modelQwenFour}{\textsc{Q4B}\xspace}
\newcommand{\modelGemmaNine}{\textsc{G9B}\xspace}

\makeatletter
\DeclareRobustCommand\onedot{\futurelet\@let@token\@onedot}
\def\@onedot{\ifx\@let@token.\else.\null\fi\xspace}

\makeatother

\DeclarePairedDelimiter{\ceil}{\lceil}{\rceil}

\ifsubmit
\else

\usepackage{fancyhdr}
\usepackage{datetime}
\renewcommand{\headrulewidth}{0pt}
\renewcommand{\footrulewidth}{0.4pt}

\fi

\title{
Rethinking Easy-to-Hard: Limits of Curriculum Learning \\ in Post-Training for Deductive Reasoning
}

\newcommand{\mpi}{1}
\newcommand{\ethz}{2}
\newcommand{\hk}{3}

\author{\textbf{Maximilian Mordig}$^{\mpi,\ethz}$~\;~\;~\textbf{Andreas Opedal}$^{\mpi,\ethz}$~\;~\;\textbf{Weiyang Liu}$^{\mpi,\hk}$~\;~\;~\textbf{Bernhard Schölkopf}$^{\mpi,\ethz}$ \\
$^{\mpi}$Max Planck Institute for Intelligent Systems, T{\"u}bingen \\
$^{\ethz}$ETH Z{\"u}rich
    \quad $^{\hk}$The Chinese University of Hong Kong\\
    \texttt{\href{maximilian.mordig@tuebingen.mpg.de}{maximilian.mordig@tuebingen.mpg.de}} 
}

\begin{document}

\maketitle

\begin{abstract}
Curriculum learning (CL), motivated by the intuition that learning in increasing order of difficulty should ease generalization, is commonly adopted both in pre-training and post-training of large language models (LLMs). The intuition of CL is particularly compelling for compositional reasoning, where complex problems are built from elementary inference rules; however, the actual impact of CL on such tasks remains largely underexplored. We present a systematic empirical study of CL for post-training of LLMs, using synthetic arithmetic and logical benchmarks where difficulty is characterized by reasoning complexity rather than surface-level proxies. Surprisingly, across multiple model families and curriculum schedules, we find \emph{no} robust advantage in difficulty-based sequencing over standard random sampling in either accuracy or response length. These findings persist across both supervised fine-tuning (SFT) and reinforcement learning (RL) methods. Our study suggests that, in the context of deductive reasoning, the specific ordering of training examples plays a negligible role in achieving compositional generalization, challenging the practical utility of curriculum-based post-training.\looseness=-1
\end{abstract}
\section{Introduction}
Recent post-training approaches---notably supervised fine-tuning (SFT) with chain-of-thought traces \citep{wei2022chain, ho_etal_2023_large} and reinforcement learning (RL; \citealp{ouyang2022training, rafailov2023direct, guo2025deepseek}) with verifiable rewards (RLVR)---have significantly extended the reasoning capabilities of large language models (LLMs) beyond their initial pre-training. However, generalization from solving easy to more complex instances of reasoning problems of the same type often remains limited~\citep{dziri2023faith,kordi2025revisiting,malek2025frontierllmsstrugglesimple,shojaee2025illusion}, suggesting that models do not learn the underlying rules that govern reasoning composition.

Indeed, most reasoning tasks are inherently \emph{compositional}: solutions to harder problems can be constructed by combining solutions to simpler subproblems. This structure naturally invites \textbf{curriculum learning} (CL; \citealp{bengio2009curriculum,hacohen2019power}), a training technique in which examples are learned in a difficulty-based order, traditionally from easy to hard. CL has a long history in machine learning \citep{elman1993dn,soviany2022curriculum} and is commonly applied in large-scale LLM pre-training \citep{brown2020language,nagatsuka2021pre,li2022stability,pouransari2024dataset,zhang2025beyond}. To learn reasoning in particular, one might expect that mastering simple instances first would allow the model to internalize the rules required for generalizing to complex compositions.

Despite its intuitive appeal, the role of CL in \emph{post}-training for reasoning remains poorly understood.
While CL is sometimes integrated into the post-training pipeline for LLMs (e.g., \citealp{havrilla2024teaching,team2025kimi}), the effect of CL on reasoning performance has not been studied systematically. 
In particular, it is unclear whether ordering examples by difficulty improves generalization to harder instances, where difficulty is characterized by underlying reasoning structure rather than through some proxy of reasoning difficulty such as, e.g., number of tokens.
Since existing reasoning benchmarks are often contaminated by pre-training data~\citep{jacovi-etal-2023-stop,zhang2024carefulexaminationlargelanguage}, it can be difficult to isolate true generalization effects.

In this short paper, we conduct a controlled empirical study of CL for post-training on deductive reasoning datasets that have compositional structure. We employ synthetic arithmetic and logical datasets, where annotated difficulty is tied to reasoning complexity rather than surface-level features. This allows us to evaluate generalization to more complex problems under the same logical rules, while avoiding confounds stemming from contaminated training data. 
We train several medium-sized models using both SFT and RL with GRPO \citep{guo2025deepseek}, comparing multiple curriculum schedules---including increasing, decreasing, and mixed-difficulty variants---against standard sampling, under a fixed training budget.

Across all experimental settings, CL yields \emph{no} consistent accuracy gains over standard sampling. We also observe that response lengths---a key factor in reasoning performance~\citep{su2025between}---are largely invariant across curricula. While perhaps counterintuitive, these findings align with broader empirical studies \citep{zhang2018empirical,wu2020curricula} suggesting that no specific curriculum strategy reliably outperforms randomized training.
Moreover, we observe that RL generally improves out-of-distribution accuracy; SFT, on the other hand, can in some cases even \emph{degrade} performance relative to a zero-shot baseline.
 
Our results suggest that example ordering plays a negligible role in post-training for deductive reasoning. In particular, a systematic easy-to-hard ordering of training examples does \emph{not} seem to help the LLM learn the underlying rules of composition. We thus question the practical utility of CL for training LLM-based reasoning models, motivating further research on alternative training methods that can achieve robust compositional generalization on reasoning tasks.
\section{Related Work}

\paragraph{Post-training for reasoning.}
The most common training methods for improving reasoning in LLMs are (i) supervised fine-tuning on chain-of-thought (CoT; \citealp{wei2022chain}) traces that verbalize solution trajectories and (ii) RL methods such as GRPO and PPO with objective, ground-truth reward signals (i.e., \emph{verifiable} rewards). The rewards are typically based on the LLM's final answer to the problem \citep{shao2024deepseekmath, guo2025deepseek, yu2025dapo}; however, it has also become common to design more dense reward functions based on intermediate steps (e.g., \citealp{lightman2023letsverifystepstep,wang-etal-2024-math,zhang-etal-2025-lessons}).
Recent work suggests that RL yields better generalization than SFT, potentially avoiding some forms of overfitting to training traces~\citep{Chu2025mx}. We explore both training methods in the context of CL.\looseness=-1

\paragraph{Reasoning evaluation.}
Benchmarks such as GSM8K~\citep{cobbe2021training}, MATH~\citep{hendrycks2021measuring}, and AIME~\citep{aime24} are widely used to evaluate LLM reasoning capabilities. However, reported results might sometimes be misleading \citep{wu2025reasoning,llmrl2025incorrect,sainz2024data}, where data contamination from pre-training corpora is one factor that may complicate the interpretation of improvements stemming from post-training~\citep{wang2025reinforcement,shao2025spurious}. To mitigate these concerns, recent work has advocated the use of synthetic reasoning datasets with controllable structure and difficulty~\citep{chen2025justlogic,opedal2024mathgap}.
We adopt this paradigm in order to isolate the effects of CL on compositional generalization.

\paragraph{Curriculum learning.}
Curriculum learning (CL) is a training technique for improving optimization and generalization which presents training examples in a difficulty-based order, often from easy to hard~\citep{bengio2009curriculum}. It has been applied across several domains of machine learning with mixed empirical results~\citep{cirik2016visualizing,zhang2018empirical,wu2020curricula,soviany2022curriculum,xie2025logic}; for instance, performance might be highly sensitive to the choice of difficulty measures and curriculum schedules \citep{zhang2018empirical}.
In LLM pre- and post-training, data mixing and staged inclusion strategies resemble CL, though their utility is rarely isolated systematically~\citep{brown2020language,touvron2023llama,team2025kimi}.
While training on easy reasoning problems can help LLMs generalize to harder ones \citep{hase2024unreasonable,sun2024easy}, previous studies have not investigated the role of CL.
\citet{xie2025logic} apply CL on the same logical reasoning task we consider (\cref{sec:datasets}); however, they only presented results for a single experiment with in-distribution test data.
We perform a systematic study on CL for problems with a clear compositional structure, where, intuitively, CL should offer positive utility if LLMs are able to learn latent abstractions from easy problems.

\section{Experiments}
This section discusses our experimental setup; it explains CL (\cref{sec:curriculum-learning}), introduces the deductive reasoning datasets (\cref{sec:datasets}), and presents training and evaluation protocols (\cref{sec:training-evaluation-protocol}).

\subsection{Curriculum Learning}\label{sec:curriculum-learning} 
Let $\mathcal{X}$ denote the input space for a particular problem type.
We define a \textbf{difficulty function} $f\colon \mathcal{X} \mapsto \mathbb{N}_{>0}$, mapping each example $x \in \mathcal{X}$ to a $\mathbb{N}_{>0}$-valued score representing the difficulty of $x$, where a higher score constitutes a harder difficulty.
In this work, difficulty is defined based on the structure of the reasoning problem (see \cref{sec:datasets}), rather than surface level features based on natural language verbalizations, such as the length of the token sequence or number of sentences. 
A curriculum strategy specifies, at each \textbf{training phase} (e.g., one or more epochs), a subset of difficulty levels from which examples are sampled.
In the common easy-to-hard variant, training begins with simpler examples and progressively incorporates more difficult ones. Other variants reverse this order or vary the range of allowed difficulties per phase.

We fix the training budget, i.e., the number of gradient updates with a given batch size, across the different curriculum strategies. This ensures a fair comparison and that scheduling does not depend on design choices around performance metrics, which are sometimes used for curriculum scheduling \citep{soviany2022curriculum}. Each training phase lasts for a given, dataset-specific number of epochs and each epoch presents the same number of examples; see \cref{sec:operationalizing_curriculum_strategies} for more details.
In the main text we compare standard uniform sampling with an easy-to-hard curriculum strategy in which each phase presents examples from one difficulty level and the difficulty level increases over phases. 
Some studies suggest that it might be nontrivial to generalize from \emph{hard-to-easy} examples \citep{yang2024can,pikus2025hard}, so we explore such curriculum strategies as well; see \cref{fig:curricula_illustration} (illustration), \cref{sec:operationalizing_curriculum_strategies} (details), and \cref{sec:additional_results} (results).

\subsection{Datasets}\label{sec:datasets}
We evaluate on synthetic arithmetic and logical reasoning tasks where difficulty is defined explicitly by the structure of the reasoning process; see \cref{fig:problem_examples_proof_trees} for illustrations. The datasets allow us to directly control the difficulty through the data generation process.
We summarize the two datasets below and provide more details in \cref{sec:dataset_description}, with example problems in \cref{tab:problem_examples_bigger}. 
Our aim is to study whether mastering easier instances can help with harder ones, or conversely, whether mastering harder instances can help in learning easy ones.

\paragraph{Arithmetic reasoning.} We use MathGAP~\citep{opedal2024mathgap}, a synthetic dataset for GSM-like \citep{cobbe2021training} math word problems with annotated proof trees representing their solutions. These problems are compositional in nature because solving them requires combining several semantic and mathematical components under general rules of inference.
We focus on two types of problems present in MathGAP. For the first type (\lindepth), a problem with $n \geq 2$ axioms has $n-1$ inference steps, each with two premises and one arithmetic operation. The problems are linear in the sense that each inference step (apart from the first) takes the conclusion from a previous step as a new premise. The difficulty of a problem is defined as the number of inference steps it has. For the second type (\partwhole), a problem with $n \geq 2$ axioms has one inference step with $n$ premises and $n-1$ arithmetic operations.\footnote{We note that this is the \emph{shortest} proof; see \citet{opedal2025efficientreasoners} for more on evaluating proof efficiency. The same problem could be solved by $n-1$ binary inference steps, corresponding to an unfolding transformation on the $(n-1)$-ary rule in the underlying logic program \citep{tamaki-unfold}.\looseness=-1\label{fn:unfolding}} 
The difficulty of a problem is defined as the number of axioms it has. 
We refer to \citet[\S 3]{opedal2024mathgap} for more details.\looseness=-1

\paragraph{Logical reasoning.}
We use the synthetic knights-and-knaves (KK; \citealp{smullyan1978name}) dataset from \citet{xie2024memorization}. The difficulty of each problem is defined by its $n \geq 2$ number of characters, all of which have one of two roles: (i) a \emph{knight}, who is always truthful, or (ii) a \emph{knave}, who is always lying. The goal is to infer the true role of all characters given their claims about themselves and/or the other characters. These problems thus require composing the logical constraints imposed by the different characters' claims.\footnote{However, the compositional structure is weaker (cf., \citealp{pagin2010compositionality}) as compared to the MathGAP data, since the truth of a claim cannot be determined in isolation.} 
KK is an instance of the boolean satisfiability problem, which is famously NP-complete \citep{cook1971complexity,levin1973universal}. However, while the search space scales exponentially, short-circuit evaluation can substantially reduce the amount of search by eliminating impossible assignments early~\citep{dechter2003constraint}.
To keep the evaluation simple, we restrict our experiments to problems that have a unique solution.

\begin{figure*}[t]
  \centering
  \vspace{-8pt}
  \includegraphics[width=0.8\textwidth]{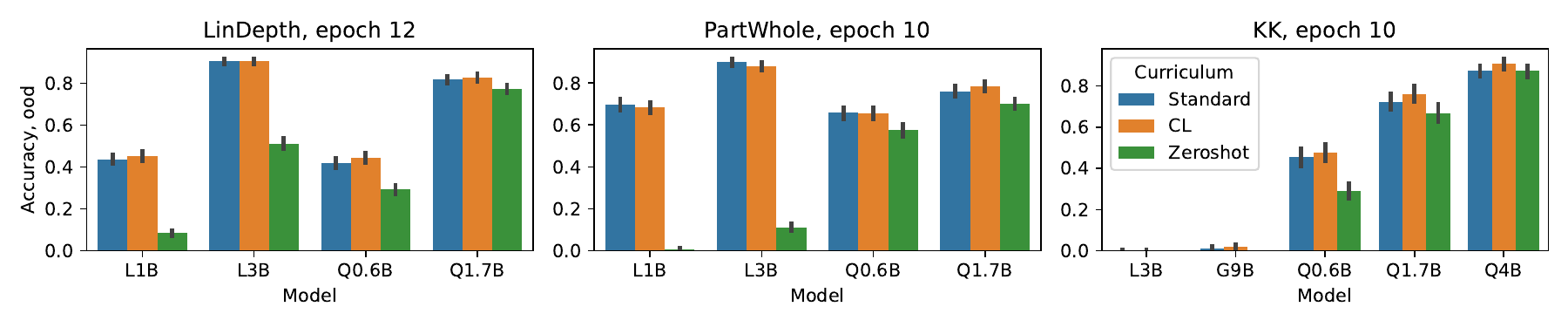}
  \includegraphics[width=0.8\textwidth]{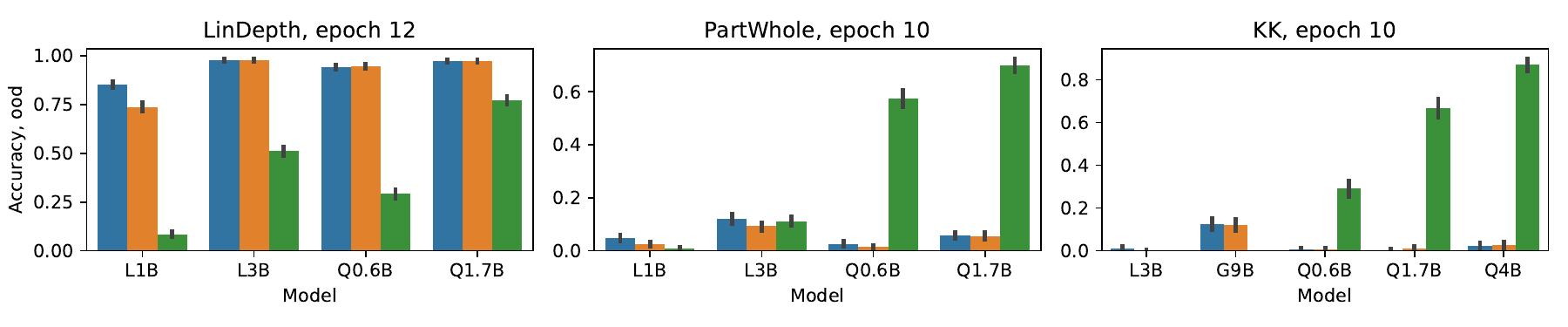}
  \vspace{-10pt}
  \caption{\ood accuracies after the final epoch for GRPO (\textbf{top}) and SFT (\textbf{bottom}) across datasets and models. We observe no consistent significant difference between standard sampling of training data and CL with an easy-to-hard curriculum strategy. \cref{fig:grpo_accuracy_metrics_extra,fig:sft_accuracy_metrics_extra} show similar results across other curriculum strategies.
  }
  \label{fig:grpo_sft_accuracy_metrics_ood_mainpaper}
\end{figure*}

\subsection{Training and Evaluation Protocol}\label{sec:training-evaluation-protocol}
We train using both SFT on annotated solution traces and RLVR with outcome-based reward signals based on correctness of the final answer and adherence to a specified format. The prompt instructs the model to ``think'' before giving the final answer, loosely motivated by \citet{kojima2022large}.
We consider both GRPO~\citep{shao2024deepseekmath} and PPO~\citep{Schulman2017ya}; the main text shows results on GRPO.
As a baseline we show zero-shot results, which suggest that model performance decreases with increasing difficulty measures for these datasets; see \cref{fig:zero_shot_accuracy}.
We choose dataset-specific token budgets based on these results as well as the annotated reasoning traces; see \cref{app:hyperparameters_and_dataset_settings}.

For each dataset, we train on a contiguous range of lower difficulty levels and evaluate on data that is \emph{in-distribution} (\indist) and \emph{out-of-distribution} (\ood), i.e., on unseen, higher difficulties.
We use, respectively, \indist and \ood difficulties 1--5 and 6--18 for \lindepth, 2--10 and 11--19 for \partwhole, and 3--6 and 7--10 for \kk.
We compare five curriculum strategies: standard uniform sampling across all training difficulties, easy-to-hard variants, hard-to-easy variants, and mixed-range variants; however, we only present results from standard sampling and one of the easy-to-hard variants in the main text since the results on the others were similar. Each strategy is trained for the same total number of gradient updates.
We experiment with the following medium-sized models:
\modelLOneF (\modelLOne; \citealp{grattafiori2024llama3herdmodels}), \modelLThreeF (\modelLThree), \modelQwenZeroSixF (\modelQwenZeroSix; \citealp{yang2025qwen3}), \modelQwenOneSevenF (\modelQwenOneSeven), \modelQwenFourF (\modelQwenFour), \modelGemmaNineF (\modelGemmaNine; \citealp{gemmateam2024gemma2improvingopen}).
We report accuracy, response length, and format compliance, all averaged over held-out test sets for different difficulty levels. 
\cref{app:implementation_details} gives implementation details and hyperparameter choices.


\section{Results}

\paragraph{Curriculum learning and performance.} \Cref{fig:grpo_sft_accuracy_metrics_ood_mainpaper}
shows the \ood~accuracies at the final epoch for GRPO and SFT.
Across datasets, models, and post-training methods, we observe no consistent gain of CL over standard uniform sampling.
This applies across all curriculum strategies (\cref{fig:grpo_accuracy_metrics_extra,fig:sft_accuracy_metrics_extra}), and differences among them are negligible, especially compared to the overall effect of post-training itself.
These findings hold across both tasks, suggesting that explicitly ordering examples by structural difficulty does not meaningfully affect compositional generalization under fixed compute. \Cref{sec:additional_results} presents additional results.

\begin{figure}[t]
  \centering
  \vspace{-10pt}
  \includegraphics[width=\columnwidth]{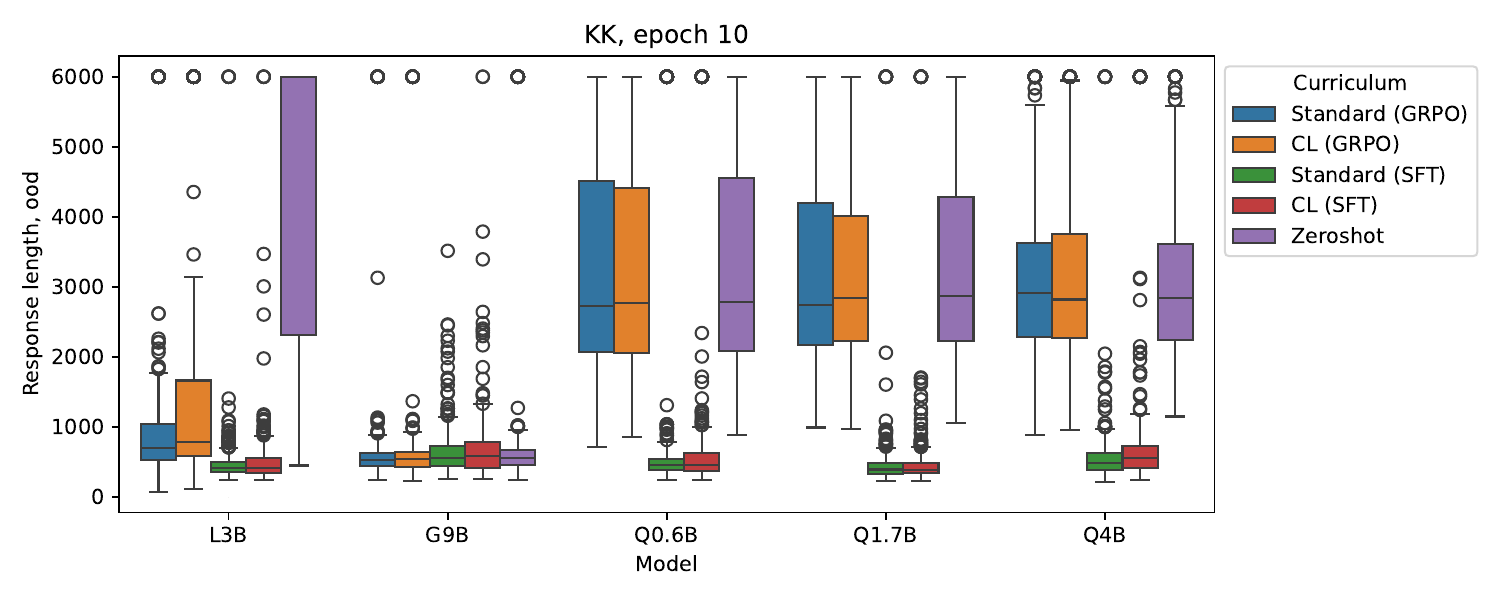}
  \vspace{-20pt}
  \caption{\ood Response lengths on \ood data after the final epoch for GRPO and SFT for the \kk dataset.
  }
  \vspace{-4pt}
  \label{fig:grpo_sft_response_lengths_metrics_ood_mainpaper}
\end{figure}

\begin{table*}
\centering
\footnotesize
\rotatebox{0}{
\begin{tabular}{llccccc|cccccc}
    \toprule
    \multicolumn{2}{c}{\textbf{\lindepth}} & $1$ & $2$ & $3$ & $4$ & $5$ & $6$ & $7$ & $8$ & $9$ & $10$ & $\geq 11$ \\\midrule
   \multirow[c]{3}{*}{\modelLOneF\!\!} & Zero-shot & 0.95 & 0.73 & 0.62 & 0.49 & 0.39 & 0.16 & 0.27 & 0.18 & 0.17 & 0.05 & 0.03 \\
    & Standard (GRPO) & 0.98 & 0.99 & 0.98 & 0.96 & 0.90 & 0.87 & 0.74 & 0.71 & 0.59 & 0.60 & 0.27 \\
    & CL (GRPO) & 0.98 & 0.98 & 0.99 & 0.95 & 0.91 & 0.86 & 0.77 & 0.73 & 0.61 & 0.63 & 0.28 \\\hline
   \multirow[c]{3}{*}{\modelLThreeF\!\!} & Zero-shot & 1.00 & 0.82 & 0.79 & 0.78 & 0.75 & 0.70 & 0.65 & 0.59 & 0.53 & 0.62 & 0.44 \\
    & Standard (GRPO) & 1.00 & 0.99 & 1.00 & 0.98 & 0.98 & 0.98 & 0.95 & 0.95 & 0.94 & 0.94 & 0.87 \\
    & CL (GRPO) & 1.00 & 0.99 & 0.99 & 0.98 & 0.97 & 0.99 & 0.96 & 0.96 & 0.94 & 0.94 & 0.87 \\\hline
   \multirow[c]{3}{*}{\modelQwenZeroSixF} & Zero-shot & 0.95 & 0.90 & 0.80 & 0.75 & 0.66 & 0.59 & 0.53 & 0.41 & 0.38 & 0.33 & 0.20 \\
    & Standard (GRPO) & 0.98 & 0.95 & 0.92 & 0.85 & 0.80 & 0.81 & 0.77 & 0.60 & 0.53 & 0.53 & 0.28 \\
    & CL (GRPO) & 0.98 & 0.96 & 0.92 & 0.88 & 0.85 & 0.81 & 0.79 & 0.62 & 0.53 & 0.56 & 0.30 \\\hline
   \multirow[c]{3}{*}{\modelQwenOneSevenF} & Zero-shot & 1.00 & 0.99 & 0.97 & 0.98 & 0.98 & 0.94 & 0.95 & 0.89 & 0.88 & 0.83 & 0.69 \\
    & Standard (GRPO) & 1.00 & 1.00 & 1.00 & 0.98 & 0.98 & 0.95 & 0.96 & 0.93 & 0.92 & 0.88 & 0.75 \\
    & CL (GRPO) & 1.00 & 1.00 & 0.98 & 0.99 & 0.99 & 0.96 & 0.97 & 0.92 & 0.94 & 0.88 & 0.76 \\\midrule
    \multicolumn{2}{c}{\textbf{\partwhole}} & $\leq 6$ & 7 & 8 & 9 & 10 & 11 & 12 & 13 & 14 & 15 & $\geq 16$ \\\midrule
   \multirow[c]{3}{*}{\modelLOneF\!\!} & Zero-shot & 0.59 & 0.10 & 0.09 & 0.05 & 0.04 & 0.02 & 0.02 & 0.00 & 0.01 & 0.01 & 0.00 \\
    & Standard (GRPO) & 0.97 & 0.97 & 0.98 & 0.97 & 0.94 & 0.93 & 0.82 & 0.85 & 0.84 & 0.69 & 0.53 \\
    & CL (GRPO) & 0.97 & 0.98 & 0.96 & 0.95 & 0.94 & 0.88 & 0.85 & 0.84 & 0.83 & 0.69 & 0.52 \\\hline
   \multirow[c]{3}{*}{\modelLThreeF\!\!} & Zero-shot & 0.97 & 0.74 & 0.55 & 0.31 & 0.26 & 0.15 & 0.09 & 0.11 & 0.16 & 0.12 & 0.10 \\
    & Standard (GRPO) & 1.00 & 0.99 & 0.98 & 0.99 & 0.95 & 0.94 & 0.93 & 0.94 & 0.95 & 0.90 & 0.87 \\
    & CL (GRPO) & 0.99 & 0.99 & 0.97 & 0.99 & 0.97 & 0.95 & 0.91 & 0.92 & 0.89 & 0.87 & 0.85 \\\hline
   \multirow[c]{3}{*}{\modelQwenZeroSixF} & Zero-shot & 0.97 & 0.93 & 0.81 & 0.77 & 0.77 & 0.70 & 0.59 & 0.64 & 0.65 & 0.57 & 0.51 \\
    & Standard (GRPO) & 0.99 & 0.95 & 0.91 & 0.88 & 0.80 & 0.79 & 0.74 & 0.66 & 0.65 & 0.70 & 0.59 \\
    & CL (GRPO) & 0.99 & 0.95 & 0.94 & 0.86 & 0.84 & 0.81 & 0.82 & 0.65 & 0.66 & 0.64 & 0.58 \\\hline
   \multirow[c]{3}{*}{\modelQwenOneSevenF} & Zero-shot & 1.00 & 1.00 & 0.98 & 0.97 & 0.93 & 0.88 & 0.88 & 0.84 & 0.71 & 0.68 & 0.58 \\
    & Standard (GRPO) & 1.00 & 1.00 & 0.99 & 1.00 & 0.89 & 0.91 & 0.91 & 0.87 & 0.82 & 0.74 & 0.65 \\
    & CL (GRPO) & 1.00 & 0.98 & 1.00 & 0.97 & 0.95 & 0.93 & 0.89 & 0.88 & 0.83 & 0.79 & 0.68 \\\bottomrule
   \end{tabular}
} 
\caption{Accuracy stratified by difficulty level for the \lindepth and \partwhole datasets under GRPO finetuning, separated by ID difficulties ($1{-}5$ and $2{-}10$) and OOD difficulties ($6{-}18$ and $11{-}19$). We observe no consistent difference between standard sampling of training data and CL using an easy-to-hard curriculum strategy.}
\label{tab:grpo_sft_accuracy_lindepth}
\end{table*}

\paragraph{RL vs. SFT.}
Across datasets, RL generally improves \ood accuracy relative to the zero-shot baseline when the model exhibits some initial capability on the task. In contrast, SFT sometimes yields limited improvements and, on certain datasets (PartWhole and KK), can degrade \ood performance below the zero-shot baseline, pointing to overfitting/memorization~\citep{Chu2025mx}.
SFT trains the model to internalize specific problem structures present in the training data, making it adapt less well to more challenging instances. RL post-training, on the other hand, encourages the model to search, which may enable it to learn the underlying logical rules of the task. 
We note that the increase in performance due to RL over SFT appears independently of curriculum strategy.

\paragraph{Response lengths.} We further analyze the empirical distributions of response lengths (\cref{fig:grpo_sft_response_lengths_metrics_ood_mainpaper,fig:grpo_response_length_metrics,fig:sft_response_length_metrics}), as RL-trained models have been shown to generate overly lengthy responses \citep{chen2025think23overthinkingo1like} which further interacts with performance \citep{su2025between}. As one might expect, we find that response length increases with difficulty (unless model accuracy drops markedly), for both RL and SFT.
The distribution of response lengths is remarkably similar across curriculum strategies.
The invariance of length dynamics across curricula provides a potential explanation for the lack of accuracy differences: example ordering does not alter the model's effective reasoning depth.

\paragraph{Generalization gap.} Finally, we note that ID accuracy is consistently higher than \ood accuracy across all settings; see \cref{tab:grpo_sft_accuracy_lindepth} for results on \lindepth and \partwhole and \cref{fig:grpo_accuracy_metrics_extra,fig:sft_accuracy_metrics_extra} for further results. This suggests that LLMs struggle to learn the underlying rules for robust generalization. Again, curriculum strategies do not significantly affect this trend. \cref{tab:grpo_sft_accuracy_lindepth} further shows that accuracy steadily decreases with difficulty.

\section{Conclusion and Implications}

The persistent appeal of easy-to-hard curricula in machine learning stems largely from their success in human learning \citep{wood-etal-1976}, where mastering atomic components is a prerequisite for navigating complex compositions.
To test whether this pedagogical motivation extends to post-training for deductive reasoning, we performed a controlled empirical study of difficulty-based curriculum learning for LLMs on synthetic compositional reasoning tasks. Across multiple datasets, model families, and training paradigms (SFT and RL), we found no consistent advantage of curriculum learning over standard random sampling under a fixed training budget.
That is, our results were negative: LLMs, for which the learning mechanisms differ from those of humans, do not seem to benefit from the same form of scaffolding. Indeed, if models learn reasoning by internalizing surface-level patterns rather than the underlying logical rules, the ordering of training examples should be less relevant. Our results further imply that the feedback mechanism provided by verifiable outcome rewards is a far more useful signal for learning composition than the ordering of training examples.

Overall, our findings challenge the practical utility of curriculum learning for compositional reasoning and suggest that future research should prioritize the structural diversity of training data and the design of robust feedback mechanisms over sequencing of difficulty.

\section*{Limitations}
Our conclusions are specific to the controlled synthetic reasoning setting studied here and to post-training with SFT and GRPO-based reinforcement learning. Although synthetic datasets allow precise control over structural difficulty and avoid contamination concerns, they do not capture the linguistic variability, ambiguity, and noise present in real-world benchmarks. Curriculum learning may therefore behave differently in settings where difficulty is less tightly defined or where language complexity plays a larger role.

We focus on medium-sized models and a fixed training budget. Curriculum effects could potentially emerge at substantially larger scales, under different compute regimes, or with alternative choices for parameters such as learning rate. In particular, curriculum learning may influence convergence speed rather than final performance, which we do not consider due to our choice of fixing the number of training steps per curriculum phase.

Finally, we examine a limited family of static, difficulty-based curricula. More adaptive strategies---such as dynamically adjusting difficulty based on model performance---could produce different outcomes \citep{setlur2025e3,shi2025efficient}. Our results therefore do not rule out broader benefits of curriculum learning, but indicate that simple difficulty-based example ordering does not consistently improve post-training generalization in the examined setting.

\newpage
\bibliography{bibliography}

\clearpage


\begin{appendices}

\crefalias{section}{appendix} 
\tableofcontents

\section{Methodology}\label{sec:app_methodology}
In \cref{sec:operationalizing_curriculum_strategies}, we give more details on the curriculum schedules, including variants absent from the main text. 
In \cref{sec:dataset_description} we provide more details on the datasets used in our experiments.

\newcommand{\mD}{\mathcal{D}}

\subsection{Curriculum Strategies}\label{sec:operationalizing_curriculum_strategies}
Recall the difficulty function $f\colon \mathcal{X} \mapsto \mathbb{N}_{>0}$, where larger values of $f(x)$ for a training example $x \in \mathcal X$ corresponds to a harder difficulty. The function $f$ is domain-specific; see \cref{sec:datasets} and \cref{sec:dataset_description}.
We are given a training dataset $\mathcal D = \{x_n\}_{n=1}^N$, where for all $x_n \in \mathcal D\colon f(x_n) \in \{1, \dots, D\}$,
with $N_d$ examples for each difficulty $d \in \{1, \dots, D\}$. Now, we split training into $D$ distinct phases, each lasting a fixed amount of $M$ ``epochs''---in this case defined as a pass over $n = N/D$ examples. We train each model for $DM + R$ epochs in total, where we additionally repeat the \emph{last} phase $R \geq 0$ times beyond the $M$ epochs, allowing more steps towards convergence.
Thus, the gradient is updated on a total of $(DM+ R)n$ samples.

\begin{figure*}[t]
    \centering
    \includegraphics[width=0.8\textwidth]{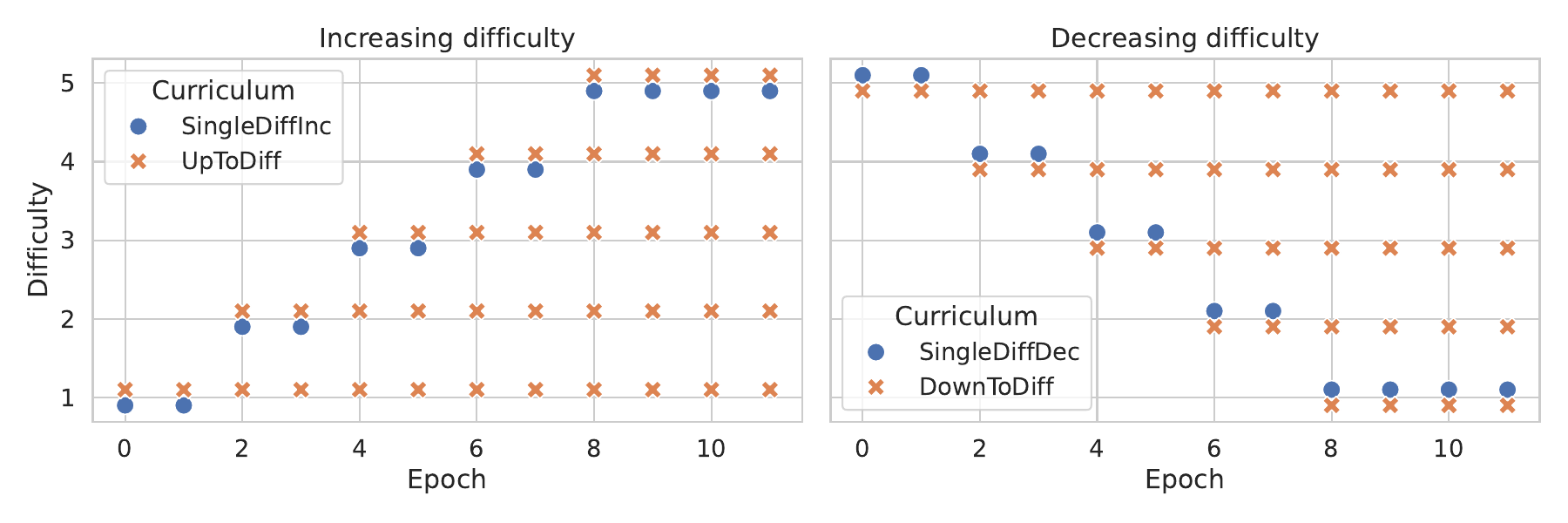}
    \caption{
    \textbf{Curriculum strategies.}
    Illustration of four of the five curriculum strategies considered in this work.
    Here, $D = 5$ difficulty levels, $M = 2$ epochs per curriculum phase, and $R = 2$ additional repetition of the final phase.
    In each epoch, $n = N/D$ datapoints are sampled from a dataset of total size $N$, with equal weight assigned to each permitted difficulty level.
    The \noCurr (uniform sampling) strategy is not shown.
    }
    \label{fig:curricula_illustration}
\end{figure*}

We consider the following five curriculum strategies (where $\ceil{\cdot}$ denotes the ceiling operator):
\begin{itemize}
    \item \noCurr~(\emph{baseline}): Standard training setting. Each epoch samples $n$ datapoints uniformly with replacement across all difficulty levels.
    \item \singleDiffInc: At epoch $i=1,\dots, DM+R$, sample $n$ datapoints from difficulty $d$, where $d=\min(\ceil{i/M}, D)$. (This strategy is used for results presented in \cref{fig:grpo_sft_accuracy_metrics_ood_mainpaper} of the main text.)
    \item \singleDiffDec: At epoch $i=1,\dots, DM+R$, sample $n$ datapoints from difficulty $d$, where $d=\max(D+1-\ceil{i/M}, 1)$.
    \item \upToDiff: At epoch $i=1,\dots, DM+R$, sample $n$ datapoints from difficulties $d=1,\ldots,\min(\ceil{i/M}, D)$, with uniform weight over difficulties.
    \item \downToDiff: At epoch $i=1,\dots, DM+R$, sample $n$ datapoints from difficulties $d=\max(D+1-\ceil{i/M}, 1), \ldots, D$, with uniform weight over difficulties.
\end{itemize}
The different strategies are illustrated in \cref{fig:curricula_illustration}.
We note that we consider the compute-restricted setting, fixing the total number of training steps, but not the number of FLOPs.\footnote{The number of FLOPs depends on the curriculum strategy via the length of the prompt, which in turn depends on the sample difficulty.} 
It would also be possible to train each model until the validation loss converges, as done, e.g., by~\citet{wu2020curricula} and \citet{soviany2022curriculum}. However, the results of such a setup would depend on the chosen validation metric.
We also do not consider more complex pacing functions, i.e., a dynamic number of samples per epoch, as this introduces substantial additional design choices.
The usual curriculum schedules considered by~\citet{erhan2009difficulty, bengio2009curriculum, brown2020language, raffel2020exploring} only consider variants of \upToDiff~and \singleDiffInc, but other schedules may be more effective, as argued in the literature~\citep{zhang2018empirical, cirik2016visualizing, soviany2022curriculum}, reflected by the \singleDiffDec~and \downToDiff~strategies.

\subsection{Datasets}\label{sec:dataset_description}
\begin{figure*}[t]
    \centering
      \includegraphics[width=\linewidth]{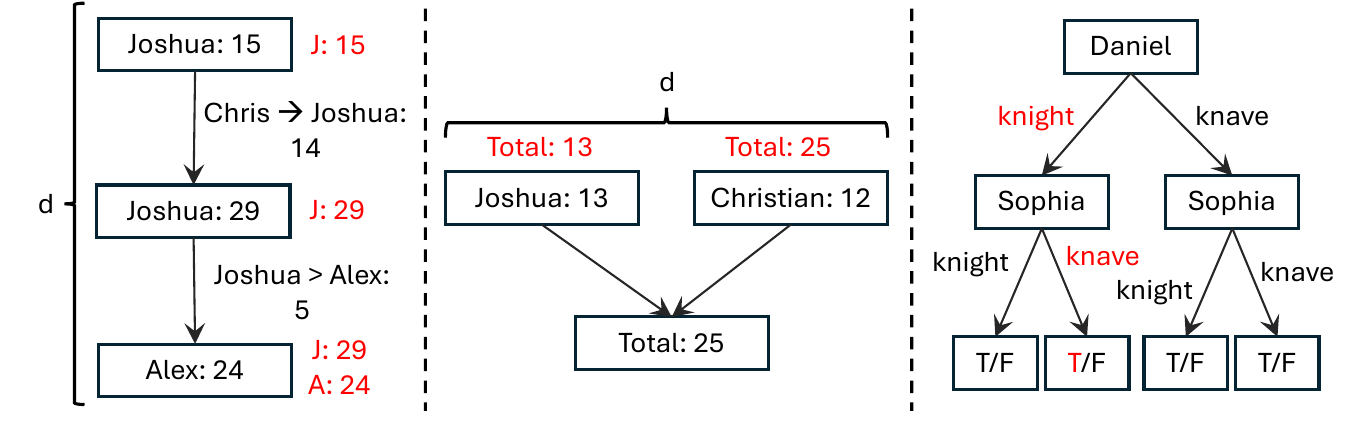}
      \caption{
        \textbf{Proof tree examples.}
        Proof trees corresponding to the low difficulty examples in \cref{tab:problem_examples_bigger}. 
        For \lindepth (left) and \partwhole (middle), red text highlights the intermediate quantities that can be tracked to solve the problem iteratively. 
        For \kk (right), the full tree is shown, with the final solution marked in red. 
        For larger problem instances, branches of the search space can often be pruned early.
      }
      \label{fig:problem_examples_proof_trees}
  \end{figure*}

We give more details on the synthetic datasets used in our study, along with the difficulty levels employed to order datapoints for CL.
All datapoints are accompanied by annotated chain-of-thought (CoT) reasoning traces.
\Cref{tab:problem_examples_bigger,fig:problem_examples_proof_trees} present example problems and informal tree representations of their underlying proofs.
In such trees, each node corresponds to an individual reasoning step and edges encode dependencies between steps.
They expose the compositional structure present in these problems, as proofs for subtrees can be combined to form proofs for larger trees.

\begin{table*}[t]
    \centering
    \tiny{
        \begin{tabular}{llp{5cm}p{4.5cm}p{2cm}}
            \toprule
            Problem Type & Difficulty & Problem & Reasoning Trace & Answer \\
            \midrule
            \lindepth & 2 & Joshua owns 15 meters of rope. Joshua then receives 14 more meters of rope from Chris. Joshua has 5 meters more than Alex. How many meters of rope does Alex have? & Joshua owns 15 meters of rope. Joshua then receives 14 more meters of rope from Chris. So Joshua has 15 + 14 = 29 meters of rope. Joshua has 5 meters more than Alex. So Alex has 29 - 5 = 24 meters of rope. & 24 \\
            \addlinespace
            \partwhole & 2 & Joshua owns 13 phones. Christian has 12 phones. If everyone sums up the phones that they have, how many phones does everybody have combined? & Joshua owns 13 phones. Christian has 12 phones. If everyone sums up the phones that they have, there are 13 + 12 = 25 in total. & 25 \\
            \addlinespace
            \kk & 2 & A very special island is inhabited only by knights and knaves. Knights always tell the truth, and knaves always lie. You meet 2 inhabitants: Daniel, and Sophia. Daniel told you that if Sophia is a knight then Sophia is a knave. Sophia said that Daniel is a knight if and only if Sophia is a knight. So who is a knight and who is a knave? & 
            Assume Daniel is a knight. No contradiction is found in their claim that If Sophia is a knight then Sophia is a knave.
            Sophia cannot be a knight, because this would contradict the claim of Daniel that If Sophia is a knight then Sophia is a knave.
            Assume Sophia is a knave. No contradiction is found in their false claim that Daniel is a knight if and only if Sophia is a knight.
            & Daniel is a knight, and Sophia is a knave. \\
            \addlinespace
            \midrule
            \lindepth & 5 & Alexander has 20 liters of milk. Alexander has 16 liters of milk more than Matthew. Olivia has 19 liters of milk more than Matthew. Ava has 12 liters of milk fewer than Olivia. Ava then receives 10 more liters of milk from Abigail. Emma has 4 liters of milk more than Ava. How many liters of milk does Emma have? & Alexander has 20 liters of milk. Alexander has 16 liters of milk more than Matthew. So Matthew has 20 - 16 = 4 liters of milk. Olivia has 19 liters of milk more than Matthew. So Olivia has 4 + 19 = 23 liters of milk. Ava has 12 liters of milk fewer than Olivia. So Ava has 23 - 12 = 11 liters of milk. Ava then receives 10 more liters of milk from Abigail. So Ava has 11 + 10 = 21 liters of milk. Emma has 4 liters of milk more than Ava. So Emma has 21 + 4 = 25 liters of milk. & 25 \\
            \addlinespace
            \partwhole & 5 & Emma has 2 vases. Ava has 5 vases. Abigail possesses 7 vases. Olivia possesses 4 vases. Matthew has 6 vases. If everyone sums up the vases that they have, what is the number of vases that everybody has in total? & Emma has 2 vases. Ava has 5 vases. Abigail possesses 7 vases. Olivia possesses 4 vases. Matthew has 6 vases. If everyone sums up the vases that they have, there are 2 + 5 + 7 + 4 + 6 = 24 in total. & 24 \\
            \addlinespace
            \kk & 3 & A very special island is inhabited only by knights and knaves. Knights always tell the truth, and knaves always lie. You meet 3 inhabitants: Henry, Alexander, and Samuel. In Henry's words: "Samuel is not a knight". As Alexander put it, "Henry is a knave". Samuel told you that If Alexander is a knave then Henry is a knight. So who is a knight and who is a knave? & Let's think step by step, by considering whether each person is lying and if that leads to contradiction. Assume Henry is a knight. No contradiction is found in their claim that Samuel is not a knight.
            Samuel cannot be a knight, because this would contradict the claim of Henry that Samuel is not a knight.
            Samuel cannot be a knave, because this would contradict the false claim of their own that If Alexander is a knave then Henry is a knight.
            We have exhausted all possibilities for Samuel, so let us go back and reconsider Henry.
            Assume Henry is a knave. No contradiction is found in their false claim that Samuel is not a knight.
            Assume Samuel is a knight. No contradiction is found in their claim that If Alexander is a knave then Henry is a knight.
            Assume Alexander is a knight. No contradiction is found in their claim that Henry is a knave. This leads to a feasible solution. & Henry is a knave, Alexander is a knight, and Samuel is a knight. \\
            \addlinespace
            \bottomrule
        \end{tabular}
    }
    \caption{
    \textbf{Dataset examples.}
    For each problem type, the problem description, ground-truth reasoning trace, and final answer are shown for very low (top rows) and medium difficulty levels (bottom rows).
    For all problem types, the number of sentences in the problem description scales approximately linearly with difficulty (up to an additive offset).
    \kk is a Boolean SAT problem, which in principle requires exponential time to explore all possible solutions,
    but many possibilities can be pruned early (see \cref{fig:ground_truth_token_lens_per_diff}).
    }
    \label{tab:problem_examples_bigger}
\end{table*}

\paragraph{Arithmetic reasoning.}

We consider two subtypes of math word problems derived from \mathgap \citep{opedal2024mathgap}.

For a \lindepth problem with $n$ axioms, there are $n$ characters who possess some quantity of some entity object. There are two arithmetic concepts present in these problems: the characters may either \emph{transfer} integer quantities of objects among each other or \emph{compare} the quantities of the objects they possess among each other.
In each new axiom, a new character is introduced along with such a relationship to an individual mentioned in the previous axiom. The question asks about the number of quantities that the character who was last introduced possesses; a problem with $n$ axioms therefore requires $n-1$ inference steps to solve, defining the problem's difficulty.
The reasoning structure can be represented as a proof tree with height $n-1$; see \cref{fig:problem_examples_proof_trees}.
These problems are solved by sequentially computing the quantity possessed by each character, requiring memory of the previous character and simple arithmetic rules.

For a \partwhole problem with $n$ axioms, there are once again $n$ characters who possess some quantity of some entity object. However, for these problems, the quantities are all given, and the task is to compute the total amount across all characters. This can be represented as a single inference step with $n$ premises and $n-1$ addition operators, which is how the annotated ground-truth chain-of-thought (CoT) traces are verbalized. Note, however, that it could equivalently be constructed by summing the quantities incrementally one at a time; see \cref{fn:unfolding}.
Unlike \lindepth, the problem can be solved without keeping track of the quantities possessed by each character by simply summing all integers given in the problem.

\paragraph{Logical reasoning.} 
The \textsc{Knights and Knaves} (\kk) dataset~\citep{xie2024memorization} consists of logical puzzles with $n$ characters. Each character is either a knight (which always tells the truth) or a knave (which always lies). The goal is to infer the truthfulness of all characters by analyzing the logical consistency of their statements about one another. 
This can be formulated as a boolean satisfiability problem (SAT) with possibly multiple solutions.
However, we restrict ourselves to instances that admit a unique solution.
The CoT trace iteratively constructs a role assignment, ruling out impossible assignments by following the logical consistency of the statements, and backtracking when necessary.
This task requires counterfactual reasoning, carefully ruling out impossible assignments through a series of backtracking steps.
The problem can be solved by assuming that the first character is a knight, then solving the problem for the remaining $n-1$ characters. If no contradiction is found, the first character is a knight. Otherwise, the first character is a knave. 

\paragraph{Motivating curricula through compositionality.}
Each problem has an underlying difficulty $d$---see \cref{fig:problem_examples_proof_trees}---which is used to order the problems for CL.
Problems of different difficulty levels are structurally related: mastering easier instances can help in solving harder ones, and vice versa.
For \lindepth, consider a problem $P$ consisting of $n$ axioms verbalized as sentences $s_1, \dots, s_n$, i.e., with difficulty $d=n-1$.
Assume the model has learned to solve the subproblem $P'$ consisting of the first $n-1$ axioms with sentences $s_1, \dots, s_{n-1}$. 
The problem $P$ can then be solved by first solving $P'$ and subsequently using its result together with the last axiom in sentence $s_n$ as premises in a single remaining proof step. This motivates the \singleDiffInc and \upToDiff curricula.
Conversely, if the model has learned to solve $P$, it is only required to decompose the structure of $P$ in order to learn how to solve $P'$. This motivates the \downToDiff and \singleDiffDec curricula.
For \kk, we can make a similar argument by fixing the truthfulness of one character, then solving the smaller problem and backtracking if a contradiction is found.

\begin{figure*}
  \centering
  \includegraphics[width=0.7\textwidth]{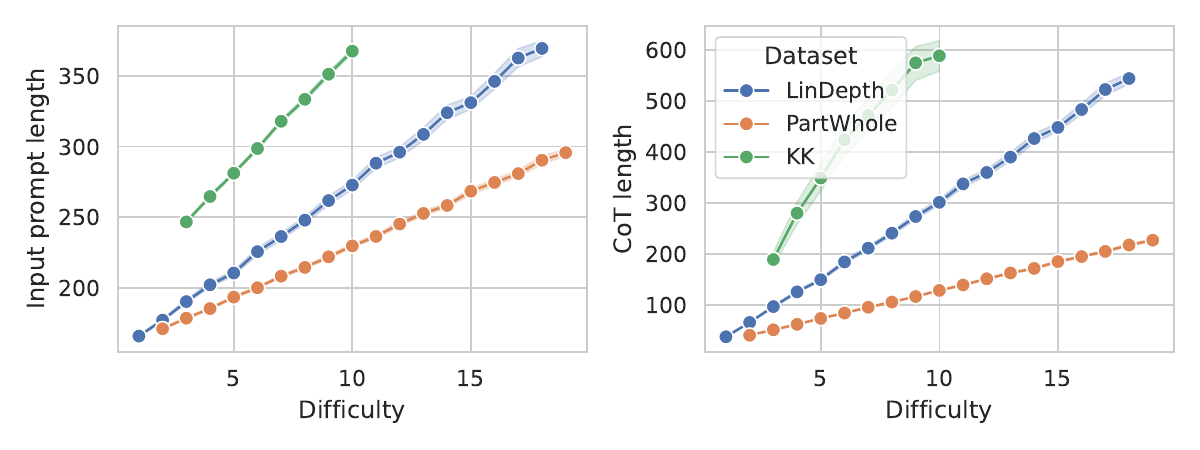}
  \caption{
  \textbf{Input prompt and ground-truth reasoning length as a function of difficulty.}
  The left panel shows the input prompt length (tokenized using the tokenizer from \modelQwenZeroSixF),
  while the right panel shows the ground-truth reasoning trace length (reasoning trace plus answer).
  Input prompt length increases approximately linearly with difficulty.
  Ground-truth reasoning length also increases linearly with difficulty for \mathgap and \partwhole,
  and sublinearly (with a steeper slope) for \kk.
  These trends inform the choice of maximum response length used during RL training and evaluation.
  }
  \label{fig:ground_truth_token_lens_per_diff}
\end{figure*}

\paragraph{Difficulty and model behavior.} 
\Cref{fig:ground_truth_token_lens_per_diff} shows the number of tokens of the prompt and the CoT trace.
We observe that the input prompt length increases linearly with problem difficulty.
In this work, we do not factor out the correlation between difficulty and input length, but instead analyze the overall effect of difficulty on performance.
Similarly, the CoT length grows linearly with difficulty for \mathgap and \partwhole, and sublinearly for \kk.
Although \kk has an exponential search space, short-circuit evaluation substantially reduces the search cost by eliminating impossible assignments early~\citep{dechter2003constraint}.

\section{Implementation Details}\label{app:implementation_details}
We use the \verl~RL framework for all experiments and implement a curriculum sampler on top of it, adapting the code where necessary. 
We branched off commit \verb|f0b4abaefc45573a591160896f8d544d8a34e45f| from \verl~(version \texttt{0.4.1.dev}). Due to active development of \verl, we updated the SFT training code script to commit \verb|3cc7695f4c70620ad871437037856f32182de096|.
\verl~also supports zero-shot generation and SFT training, which we use to reduce differences in model performance due to different frameworks as much as possible.
We extended the \verl~SFT training code to perform rollouts on the test dataset during training.
For performance reasons (as is the standard in \verl), we filter out examples with more than 800 input tokens and 1200 input+CoT tokens (with respect to the \QwenThreeSeries~tokenizer). This filters out a very small fraction of the examples.
\verl~outputs the generated rollouts as json files, which we then parse to extract the reasoning traces and answers.
All plots show the performance on the test dataset, error bars are computed across bootstrap samples.
We will make the code available in the final version of the paper.

\subsection{Hyperparameters and Dataset Settings}\label{app:hyperparameters_and_dataset_settings}
We use the following hyperparameters for the different problem types:
\begin{itemize}
  \item \lindepth~\citep{opedal2024mathgap}: train on difficulties 1--5, evaluate on difficulties 1--18, $M=2$ epochs per curriculum phase, 4096 training samples per difficulty, global batch size 128 for SFT, global batch size 256 for RL, rollout length of 2000 tokens.
  \item \partwhole~\citep{opedal2024mathgap}: train on difficulties 2--10,\footnote{Difficulty 1 is excluded; being only a single axiom, the LLMs often respond as if it is a trick question during zero-shot evaluation.} evaluate on difficulties 2--19, $M=1$ epoch per curriculum phase, 4096 training samples per difficulty, global batch size 128 for SFT, global batch size 256 for RL, rollout length of 2000 tokens.
  \item \kk~\citep{xie2024memorization}: train on difficulties 3--6, evaluate on difficulties 3--10, $M=2$ epochs per curriculum phase, 1024 training samples per difficulty, global batch size 32 for SFT, global batch size 256 for RL, rollout length of 6000 tokens.
\end{itemize}
Based on the ground-truth CoT length, we choose the generation lengths (for zero-shot and RL) to be sufficiently large yet favor conciseness. The zero-shot generation length further justifies this choice.
We evaluate on 128 test examples per difficulty.
We adjust the micro batch size / token lengths per GPU per model to avoid OoM errors (which does not affect the result).
All models are trained for one additional phase ($M$ additional epochs), repeating the last curriculum phase (\cref{sec:operationalizing_curriculum_strategies}).

Models are trained with \texttt{bfloat16} mixed precision.
Following learning rate sweeps (and following the typical range of learning rates used in the literature~\citealp{Hu2021fe,xie2025logic,guo2025deepseek}), we use a learning rate of \num{1e-4} for SFT and \num{1e-6} for RL, except for RL on \kk~, where we use a learning rate of \num{3e-7}.
We finetune SFT with LoRA (rank $r=32$), whereas we do full finetuning for RL (which receives less gradient updates due to a larger batch size).
For PPO, the critic model is initialized with the same parameters as the actor model and is trained with a learning rate of \num{1e-5}.
RL uses temperature $T=1$ during training rollouts.
Following common practice, rollouts on the test set use temperature $T=0$ resulting in deterministic rollouts. This ensures that the RL model rolls out the learned (greedy) policy rather than a stochastic variant. For comparability, we use the same temperature for SFT and zero-shot rollouts.

\subsection{Prompt Format}\label{app:prompt_format}
We use the following prompts for RL finetuning, SFT and zero-shot/test evaluation:
\begin{lstlisting}[breaklines,numbers=none]
# SFT
{TASK_DESCRIPTION} {problem} <assistant_start> <think> {reasoning_trace} </think> <answer> {answer} </answer>

# RL-finetuning (RFT), zero-shot, and evaluation prompt
{TASK_DESCRIPTION} {problem} <assistant_start> <think>
\end{lstlisting}

The \texttt{TASK\_DESCRIPTION} briefly describes the task and the expected output format. 
For both RL and evaluation, this description explicitly states the expected output format 
\verb|<think>.*</think><answer>.*</answer>| 
to parse answers reliably. 
The SFT prompt additionally provides the reasoning trace, which gives a stronger training signal than RL.
The RL prompt terminates with the \verb|<think>| tag to encourage the model to reflect on the problem before producing its answer. 
We found this particularly beneficial for adherence to the desired format in preliminary experiments. 
For the \kk~dataset, the RL prompt also includes a one-shot example, which we found helpful for stabilizing training.

\subsection{Output Parsing and Reward Functions}\label{sec:output_parsing_and_reward_functions}
We detail the output parsing for the evaluation and the RL reward function. Here, we combine extraction techniques and reward functions from various recent works~\citep{guo2025deepseek, xie2025logic, shao2024deepseekmath, yu2025dapo}.

The model completions are parsed into \emph{reasoning} and \emph{answer} segments. 
Parsing first splits on a conclusion pattern that separates the reasoning from the answer, chosen among 
\verb|</answer>|, \verb|Final answer:|, or \verb|</think>|, in this order. 
The thinking part is taken as all text before the first \verb|</think>| (if present), while the answer part begins at the first \verb|<answer>| tag (or defaults to the remainder if absent). 
The parsing fails if no conclusion pattern is detected, in which case the entire completion is treated as the answer. 
The thinking segment is ignored for evaluation (but encouraging the model to reflect on its answer first), and the answer segment is post-processed in a dataset-specific way, trying to match the expected answer format.

\paragraph{Reward functions.} 
For \mathgap, the reward is defined as:
\[
R = \text{parsingSuccessful} + \text{isInt} + 4 \cdot \text{isCorrect}.
\]
We set $R=-2$ if the response hits the token limit. 
That is, when parsing fails, we attempt to extract the answer using a set of heuristics. If the extracted answer is correct but represented as a floating-point value, the correctness reward is granted, whereas the additional integer-specific reward is not.

For \kk, we follow~\citet{xie2025logic} and define the reward as:
\[
R = \text{formatScore} + \text{answerScore},
\]
where $\text{formatScore}$ is $1$ if the completion has the correct format, else $-1$.
The answer has the correct format if it contains the tags \verb|</think>|, \verb|<answer>|, and \verb|</answer>| exactly once each, and in the correct order.
The answer score is $2$ if the answer is correct, else $-1.5$ if all characters appear in the answer, else $-2$.
If the completion has the wrong format or the answer is empty, we set $R=-2$.
Although the exact weighting of components may influence performance, our preliminary experiments suggest that results are robust to moderate changes in the reward design.
While RL training is affected by incorrect output format through the reward function, the correctness evaluation only needs to parse the answer and does not enforce the format otherwise.

\subsection{In-distribution (id), out-of-distribution (ood) metrics and generalization gap}
We define the \indist (in-distribution) accuracy as the average accuracy over the in-distribution difficulties, the \ood~(out-of-distribution) accuracy as the average accuracy over the out-of-distribution difficulties (calling the generalization to larger difficulties ``out-of-distribution'', as argued before). We define the \emph{generalization gap} as the difference between the \indist and \ood~accuracies (\genGapDef).

\section{Additional Results}\label{sec:additional_results}

\subsection{Zero-Shot}

\begin{figure*}[h]
  \centering
  \includegraphics[width=\textwidth]{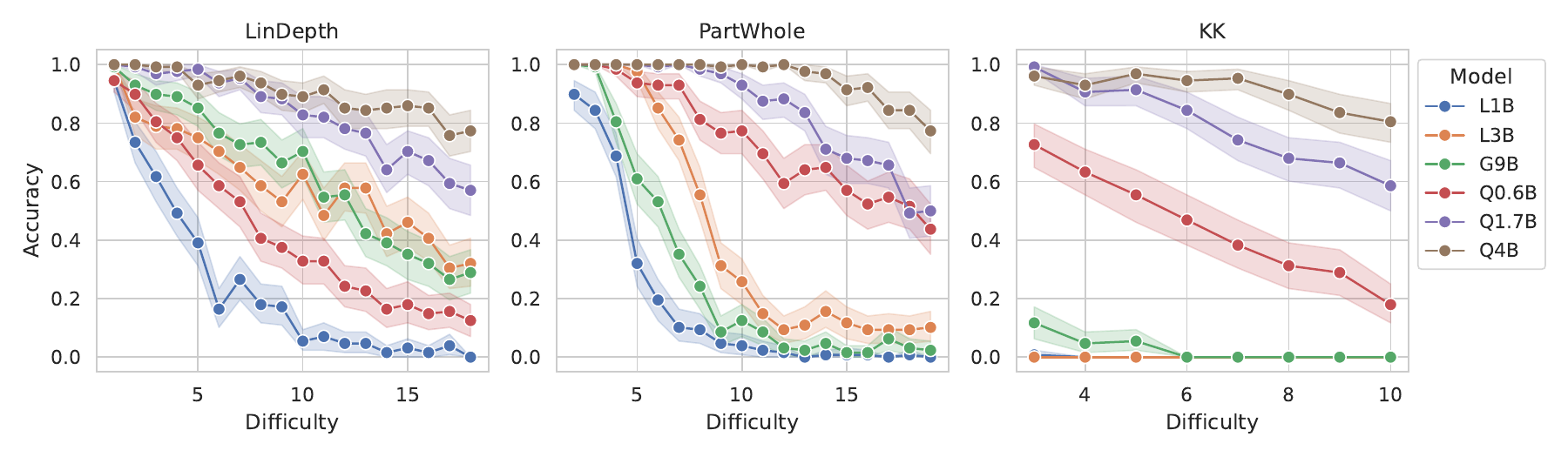}
  \caption{
  \textbf{Zero-shot accuracy across models and datasets.}
  Average accuracy is shown as a function of difficulty.
  Accuracy generally declines with increasing difficulty and drops sharply beyond the lowest difficulty levels.
  Among models with comparable size, \QwenThreeSeries models achieve the highest accuracy across all datasets.
  }
  \label{fig:zero_shot_accuracy}
\end{figure*}

\begin{figure*}[h]
  \centering
  \includegraphics[width=\textwidth]{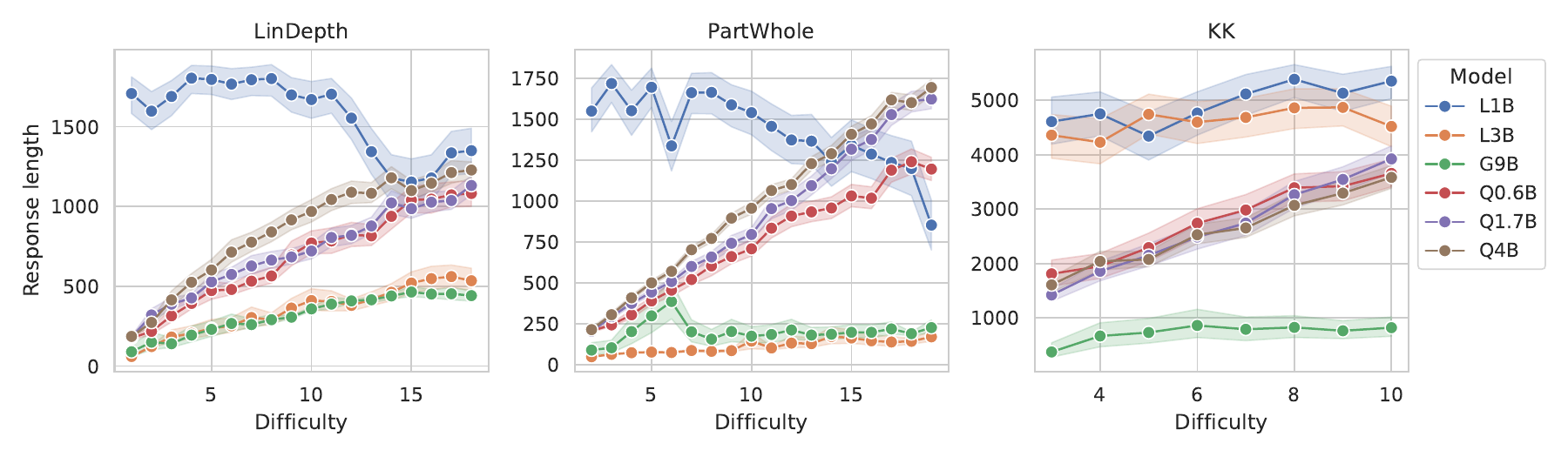}
  \includegraphics[width=\textwidth]{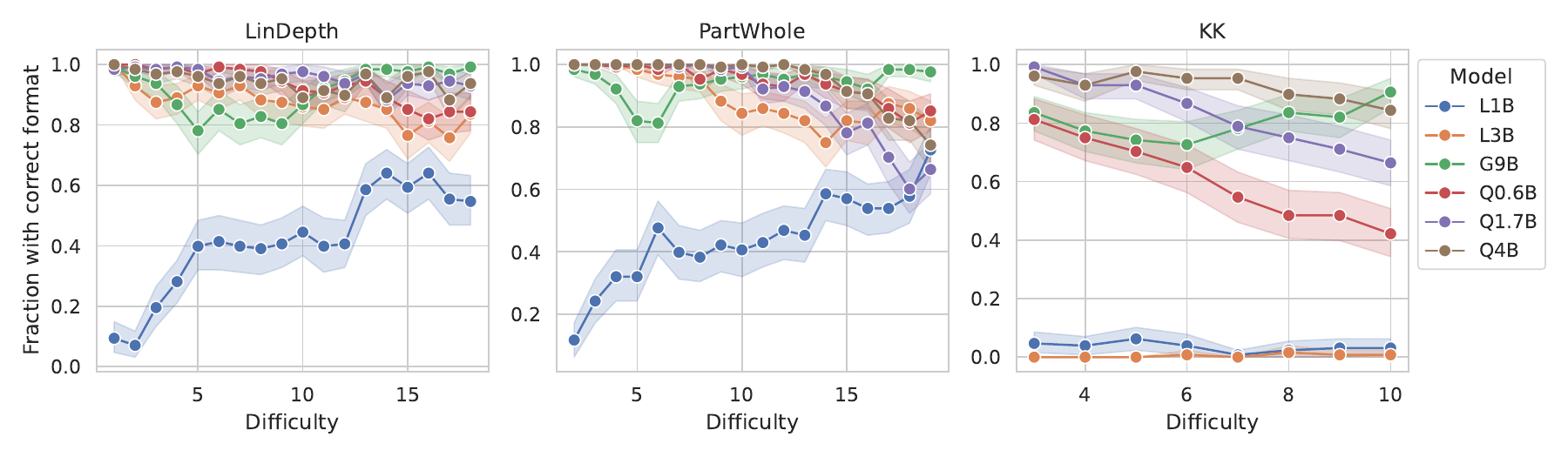}
  \caption{
  \textbf{Zero-shot response length and format correctness.}
  Response length and the fraction of correctly formatted completions are shown as a function of difficulty for each dataset.
  For most models, response length increases approximately linearly with difficulty, associated with a lower decrease in accuracy as difficulty increases (see \cref{fig:zero_shot_accuracy}).
  Except for \modelLOne, nearly all completions are correctly formatted, ensuring reliable evaluation and subsequent RL training.
  Average response lengths remain below the maximum response length used during training and evaluation.
  }
  \label{fig:zero_shot_other_metrics}
\end{figure*}

\Cref{fig:zero_shot_accuracy} shows the zero-shot performance of the models on the \lindepth, \partwhole~and \kk~datasets. \Cref{fig:zero_shot_other_metrics} shows the same for the response length and the fraction of completions that have the correct format.

Generally, the models do not exceed the allocated response length.
We observe that the defined difficulty measure is a good measure of model performance: the accuracy generally declines with increasing difficulty, and is significantly lower beyond the initial difficulties. 

\Cref{fig:zero_shot_other_metrics} shows the fraction of correctly formatted responses and response length, per dataset.

\subsection{Post-Training: Summary Metrics}\label{sec:post_training_summary_metrics}
We show summary metrics by averaging the metrics over in-distribution and out-of-distribution difficulties respectively, and show the generalization gap as the difference between the \indist and \ood~metrics.

\subsubsection{GRPO}
\begin{figure*}[h]
  \centering
  \includegraphics[width=\textwidth]{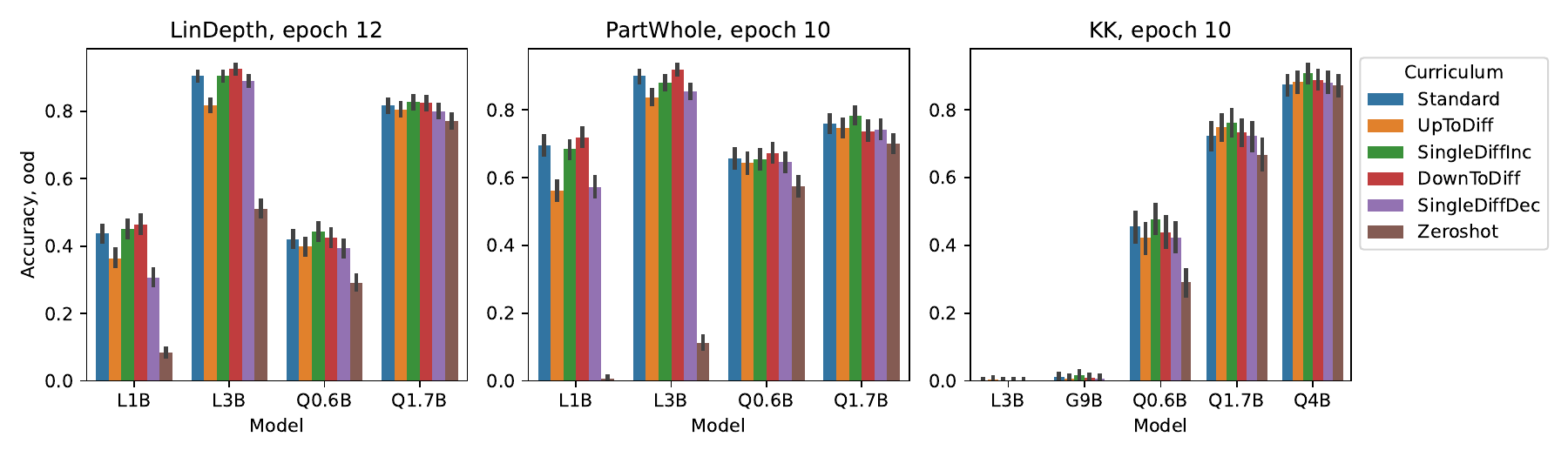}
  \includegraphics[width=\textwidth]{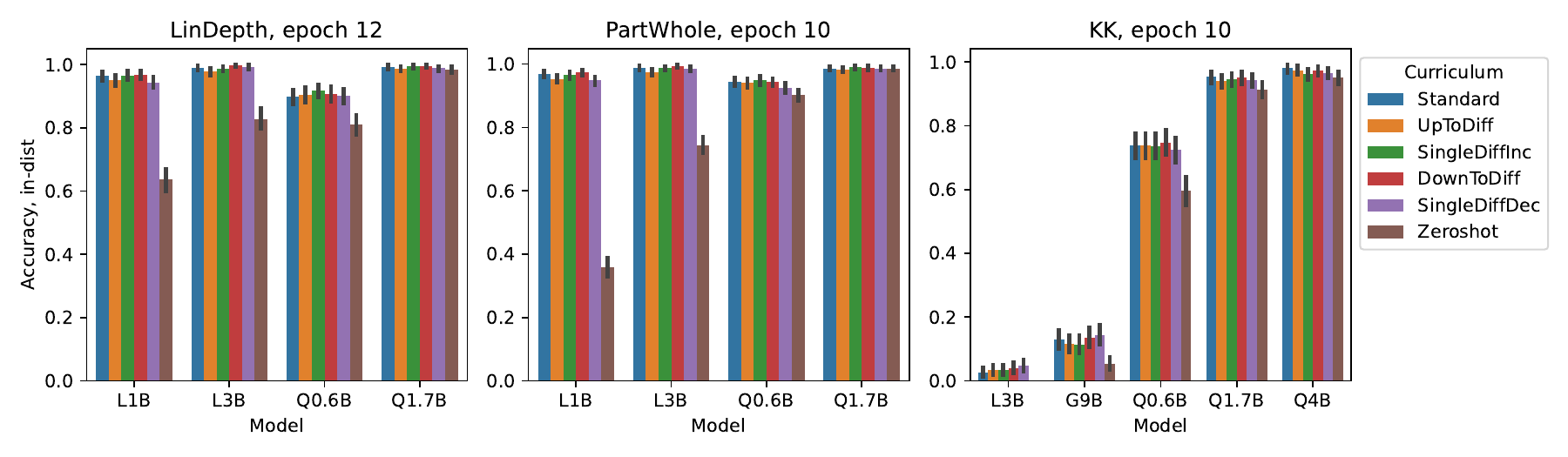}
  \includegraphics[width=\textwidth]{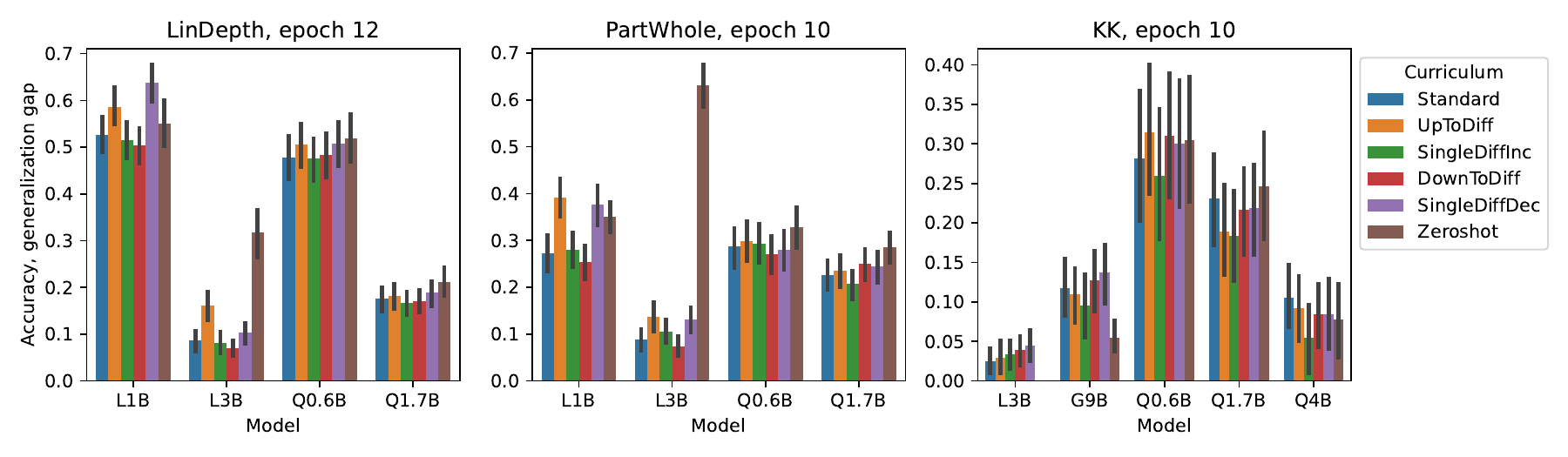}
  \caption{
  \textbf{Out-of-distribution, in-distribution accuracy and generalization gap after GRPO post-training.}
  Final-epoch \ood and \indist accuracies are shown together with the generalization gap (\genGapDef) across models and datasets.
  No curriculum strategy consistently outperforms the \noCurr~curriculum across models and datasets, including cases where post-training improves \indist performance relative to the zero-shot baseline. 
  While post-training generally reduces the generalization gap, no systematic differences are observed between curriculum strategies.
  Note the different y-axis scales, which visually enlarge the error bars for the generalization gap.
  }
  \label{fig:grpo_accuracy_metrics_extra}
\end{figure*}

\begin{figure*}[h]
  \centering
  \includegraphics[width=\textwidth]{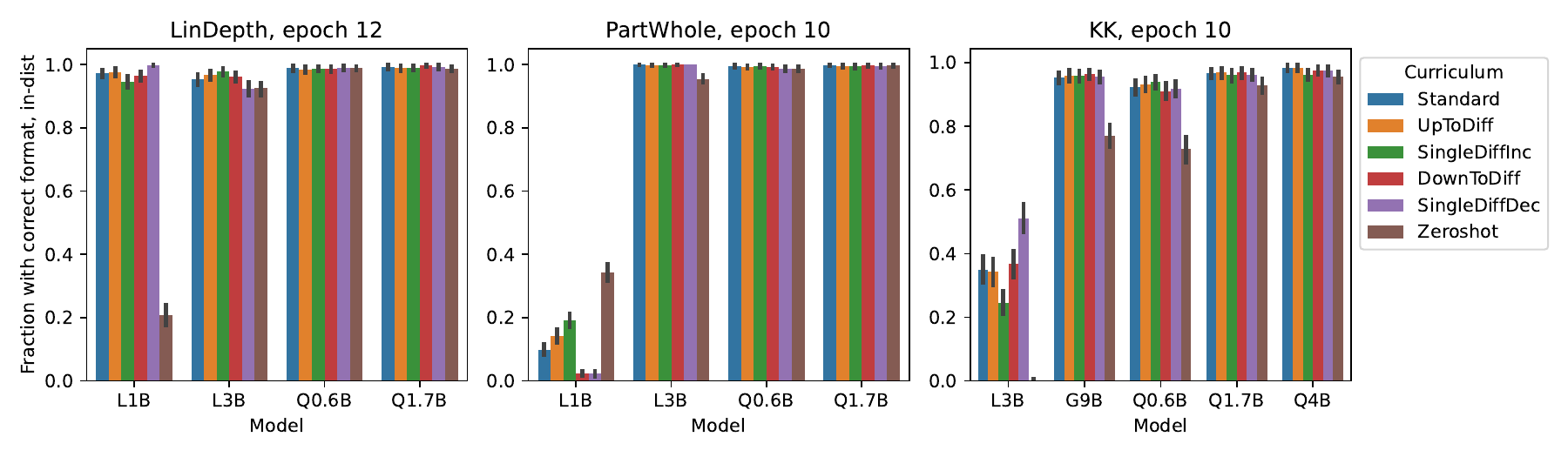}
  \includegraphics[width=\textwidth]{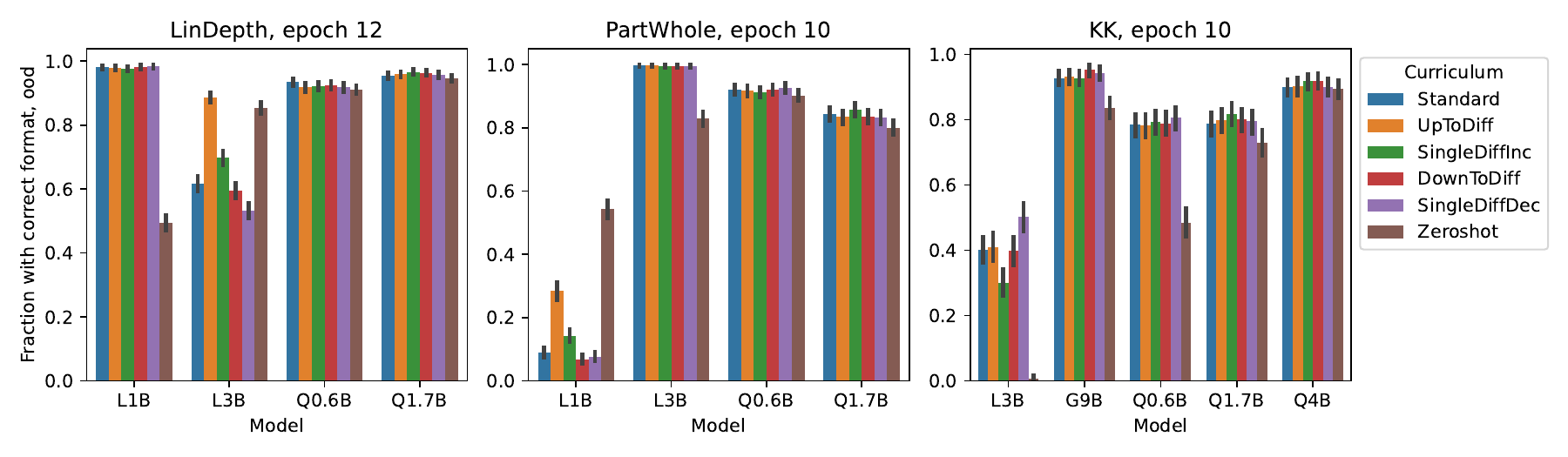}
  \includegraphics[width=\textwidth]{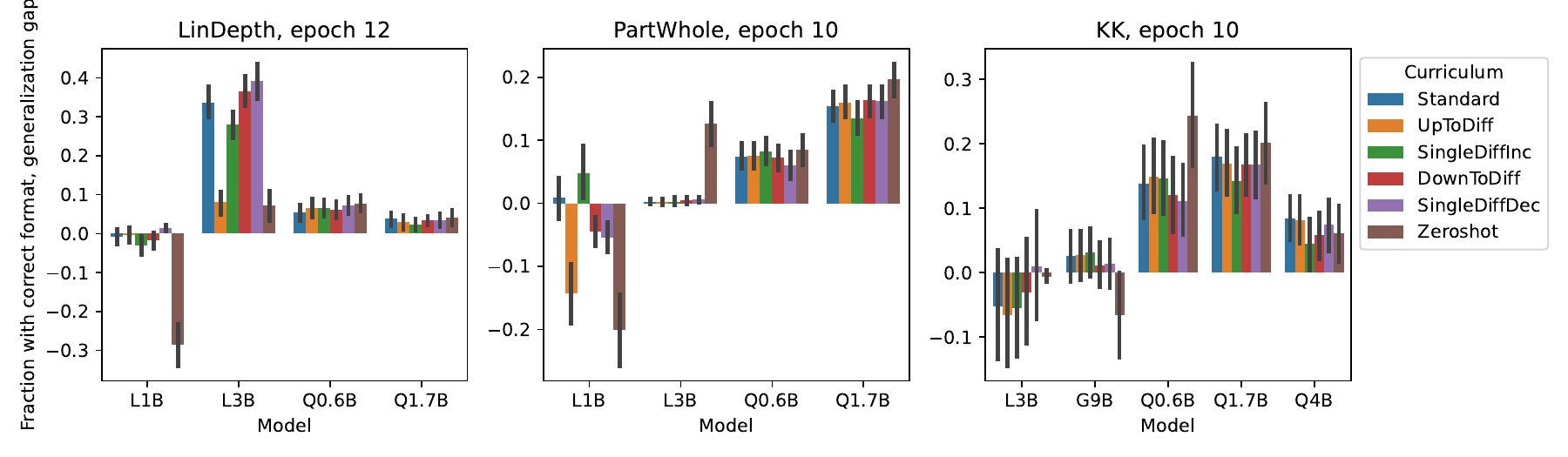}
  \caption{
  \textbf{Format adherence after GRPO post-training.}
  The fraction of correctly formatted responses is shown for in-distribution and out-of-distribution examples,
  together with the generalization gap (\genGapDef).
  Different y-axis scales are used, which visually enlarge the error bars for the generalization gap.
  }
  \label{fig:grpo_correct_format_metrics}
\end{figure*}

For GRPO, \cref{fig:grpo_accuracy_metrics_extra} shows the \ood, \indist accuracies and generalization gap at the final epoch. \Cref{fig:grpo_correct_format_metrics} repeats the same for the fraction of correctly formatted responses and response length at the final epoch.

\begin{figure*}[h]
  \centering
  \includegraphics[width=\textwidth]{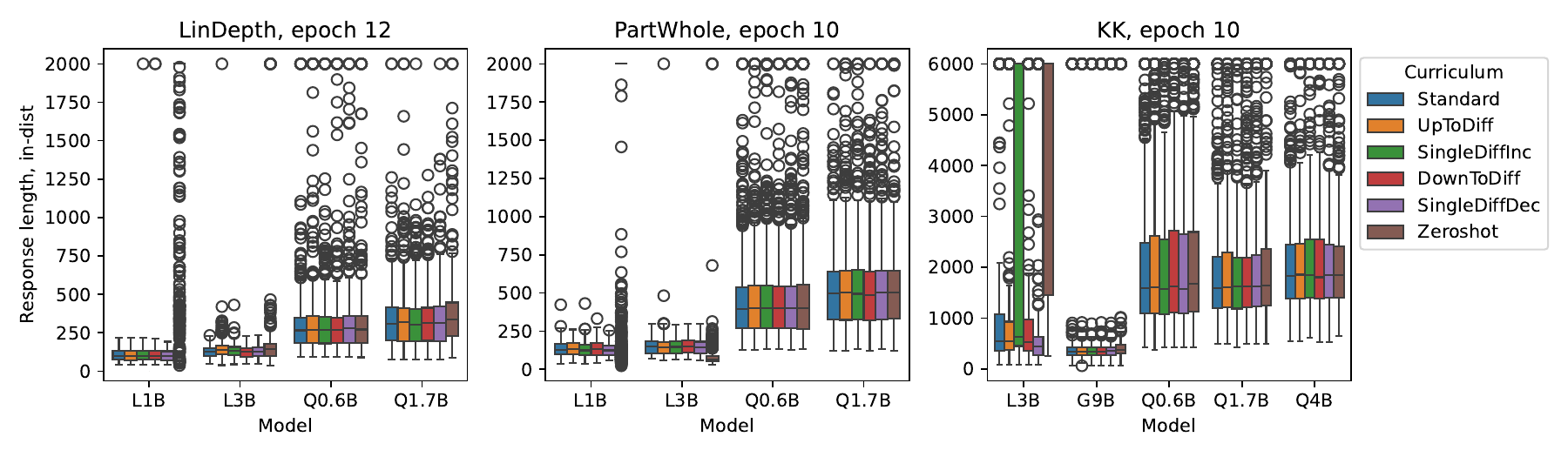}
  \includegraphics[width=\textwidth]{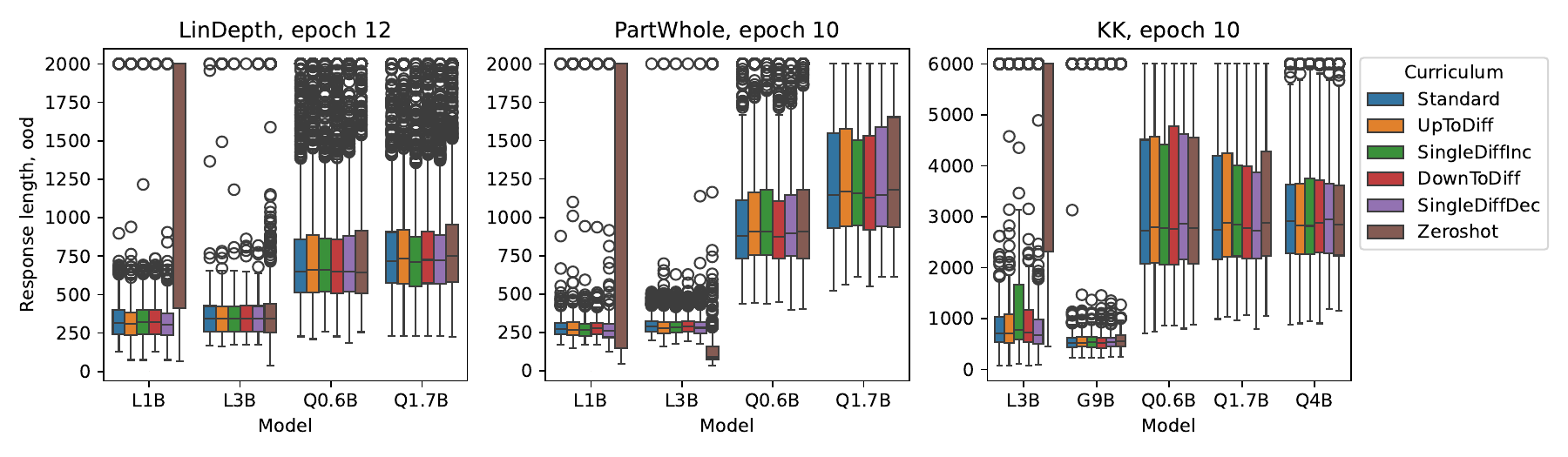}
  \includegraphics[width=\textwidth]{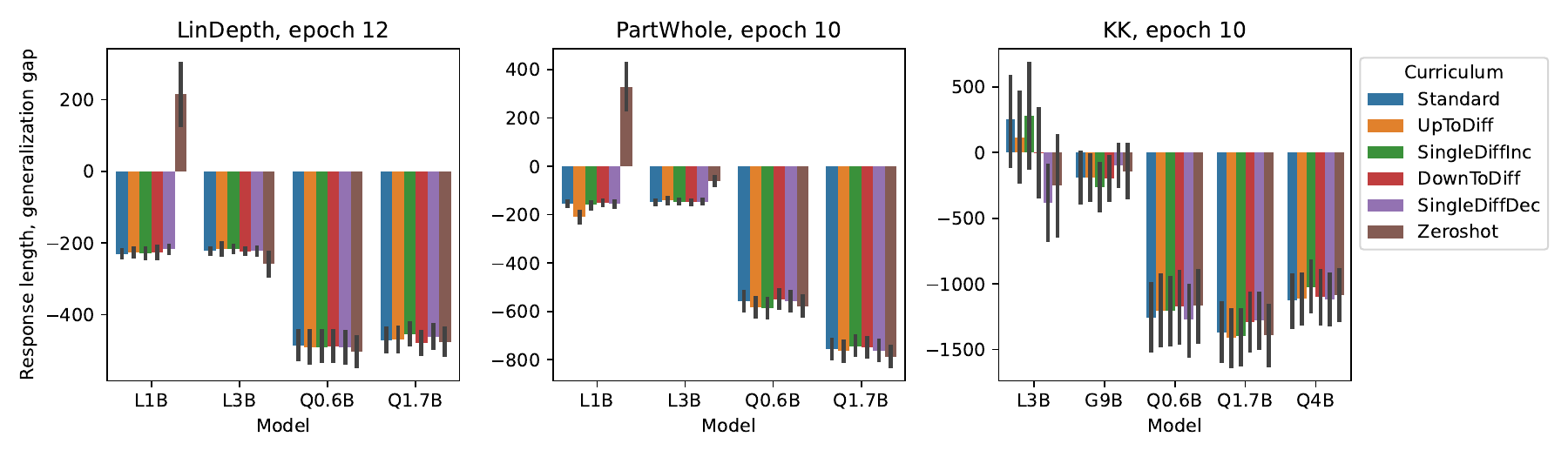}
  \caption{
  \textbf{Response length after GRPO post-training.}
  Final-epoch response lengths are shown for in-distribution (\indist) and out-of-distribution (\ood) examples,
  together with the generalization gap (\genGapDef).
  A negative generalization gap indicates that responses are longer for \ood data than for \indist data.
  }
  \label{fig:grpo_response_length_metrics}
\end{figure*}

\subsubsection{SFT}
\begin{figure*}[h]
  \centering
  \includegraphics[width=\textwidth]{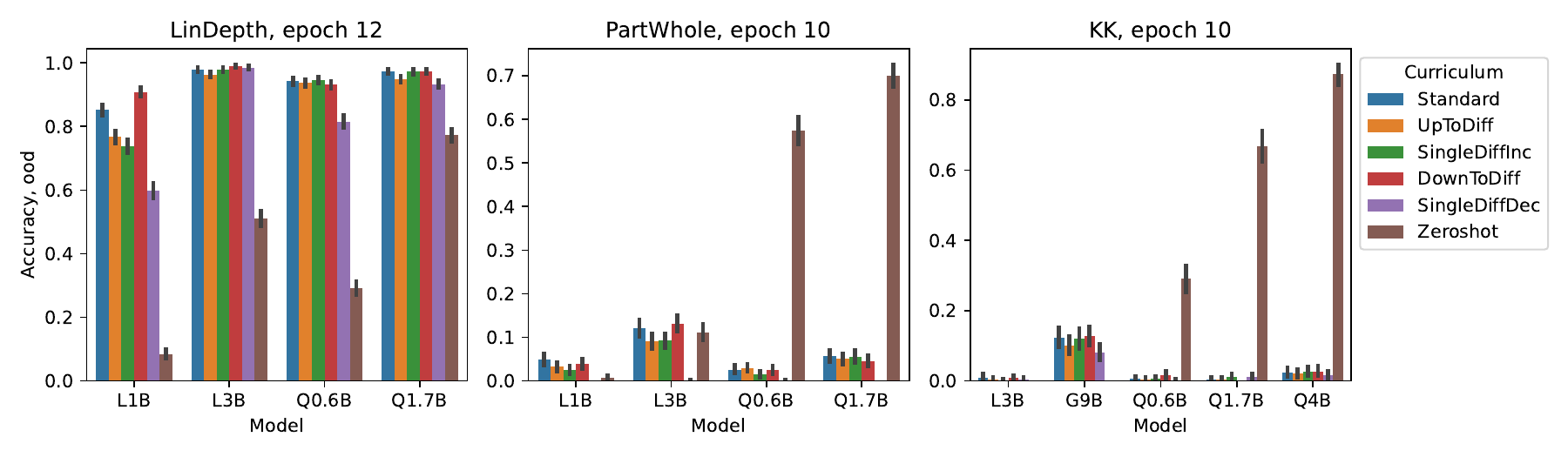}
  \includegraphics[width=\textwidth]{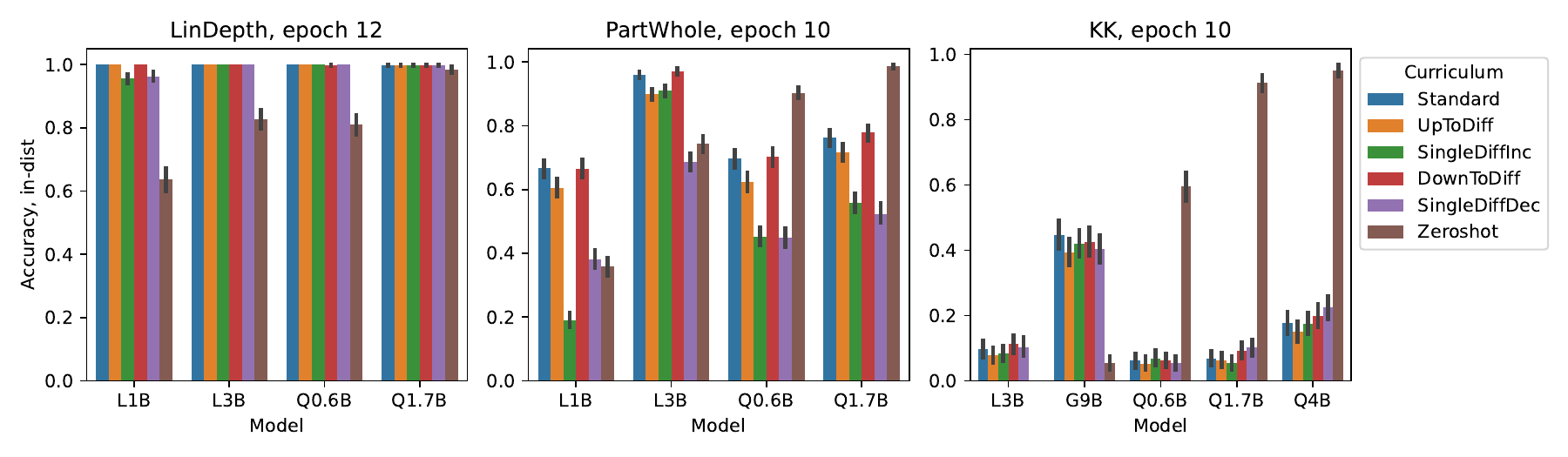}
  \includegraphics[width=\textwidth]{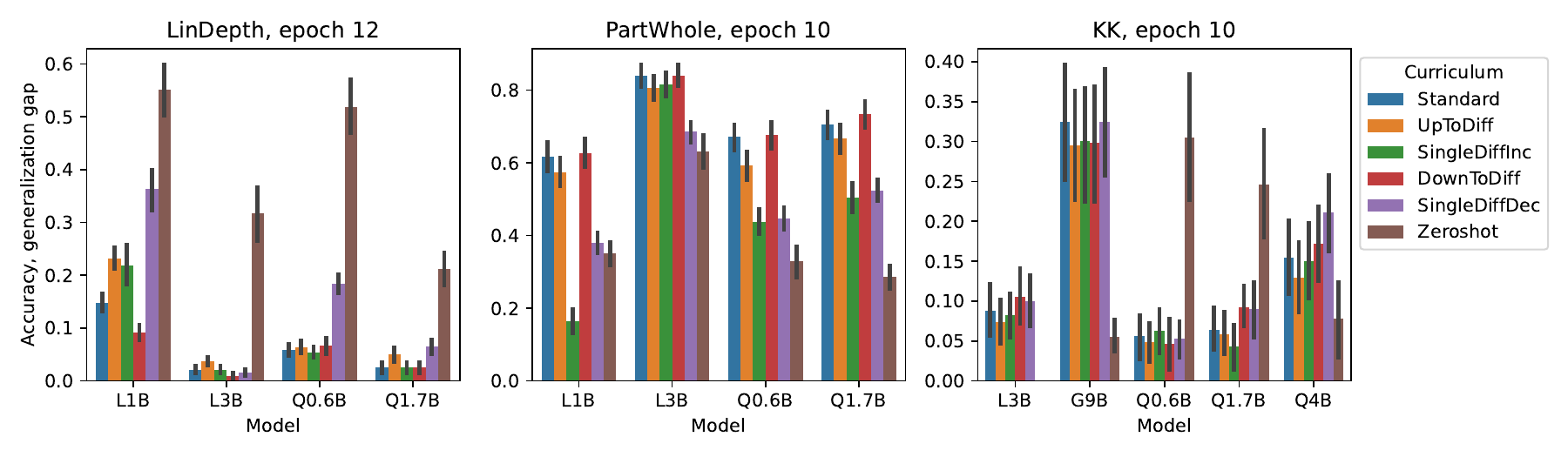}
  \caption{
  \textbf{Out-of-distribution, in-distribution accuracy and generalization gap after SFT post-training.}
  Final-epoch \ood, \indist accuracies are shown together with the generalization gap (\genGapDef).
  Note the different y-axis scales, which visually amplify the error bars for the generalization gap.
  }
  \label{fig:sft_accuracy_metrics_extra}
\end{figure*}

\begin{figure*}[h]
  \centering
  \includegraphics[width=\textwidth]{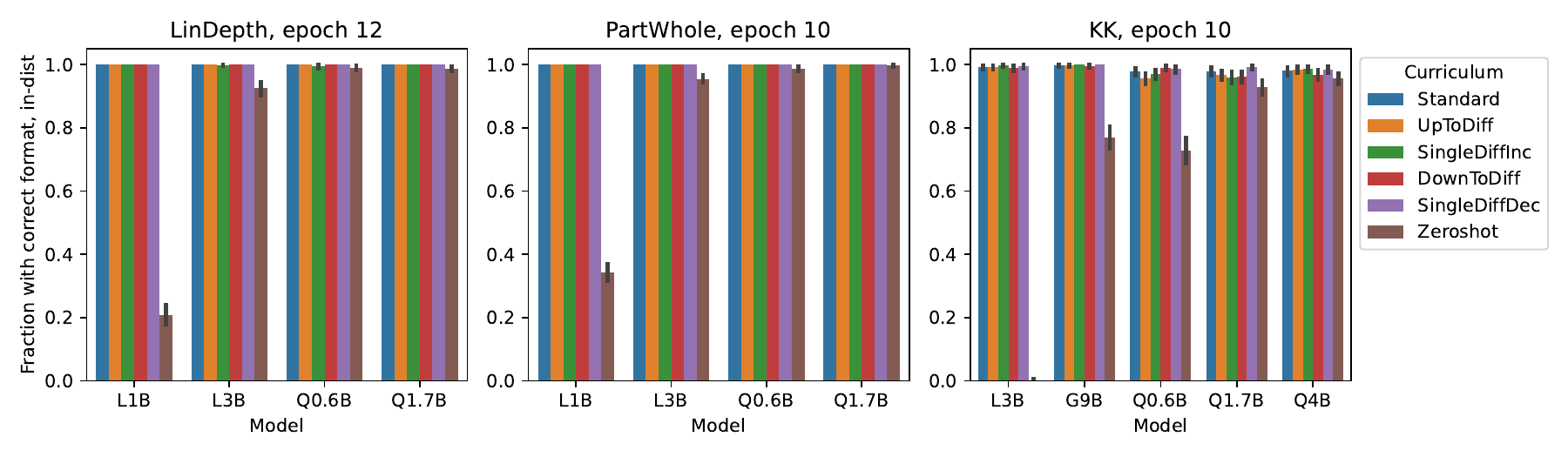}
  \includegraphics[width=\textwidth]{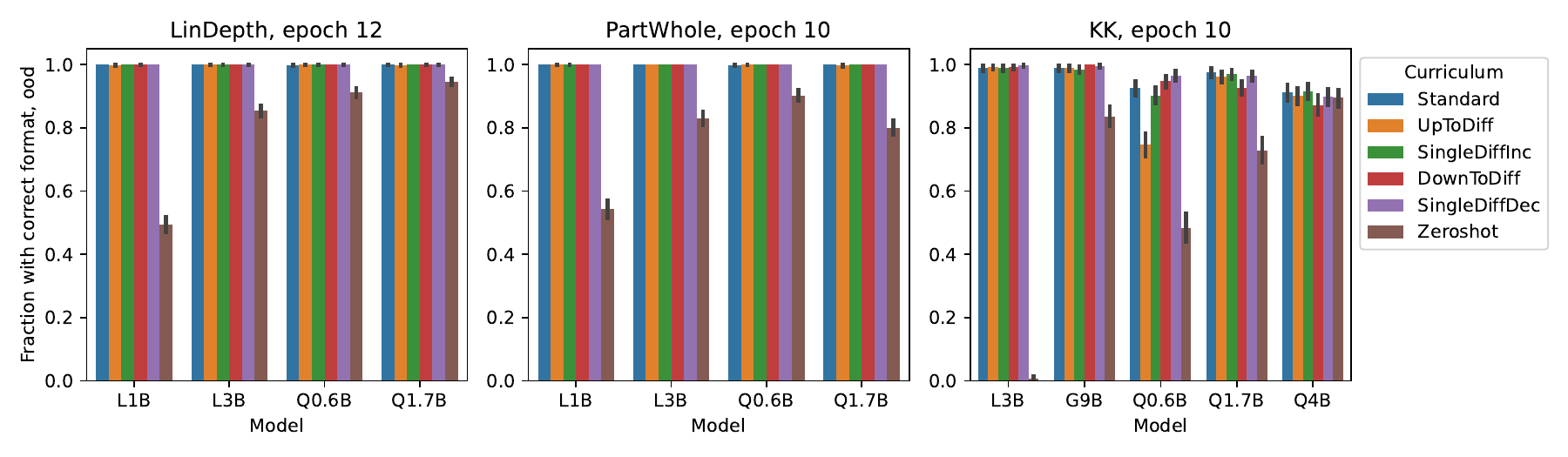}
  \includegraphics[width=\textwidth]{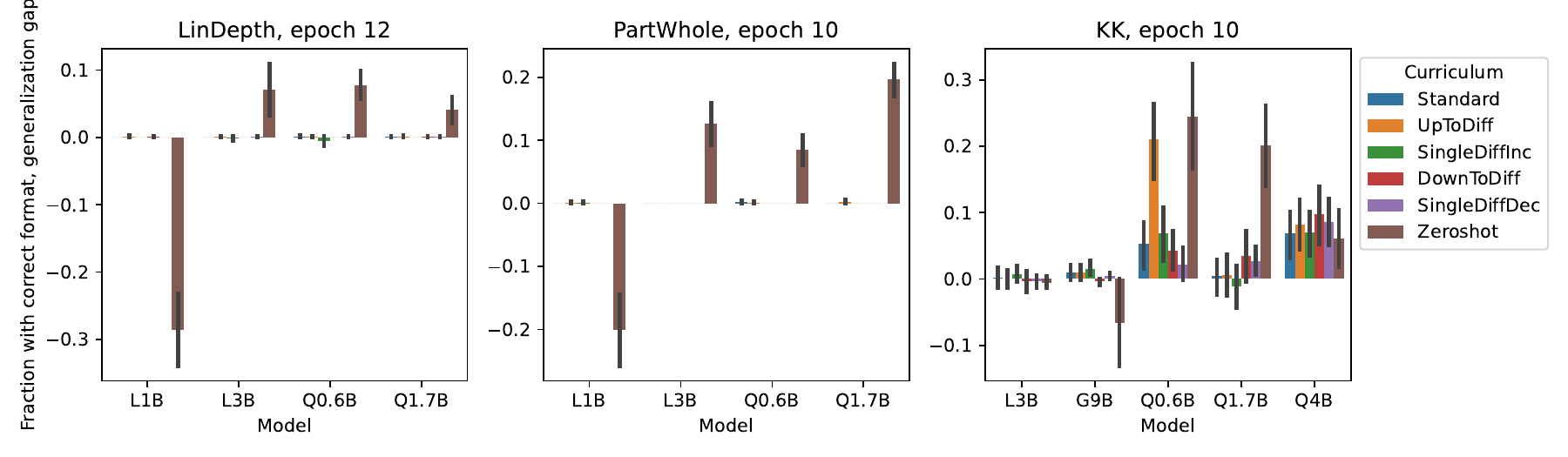}
  \caption{
  \textbf{Format adherence after SFT post-training.}
  The fraction of correctly formatted responses is shown for in-distribution and out-of-distribution examples,
  together with the generalization gap (\genGapDef).
  }
  \label{fig:sft_correct_format_metrics}
\end{figure*}

For SFT, \cref{fig:sft_accuracy_metrics_extra} shows the \ood, \indist accuracies and generalization gap at the final epoch. \Cref{fig:sft_correct_format_metrics} repeats the same for the fraction of correctly formatted responses and response length at the final epoch.
\begin{figure*}[h]
  \centering
  \includegraphics[width=\textwidth]{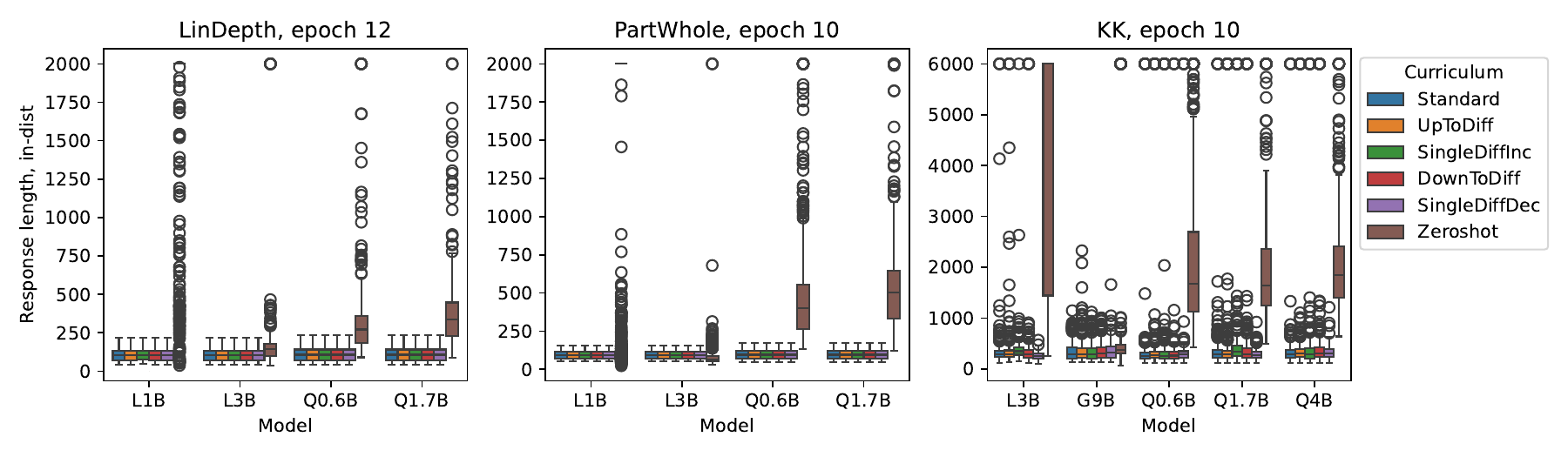}
  \includegraphics[width=\textwidth]{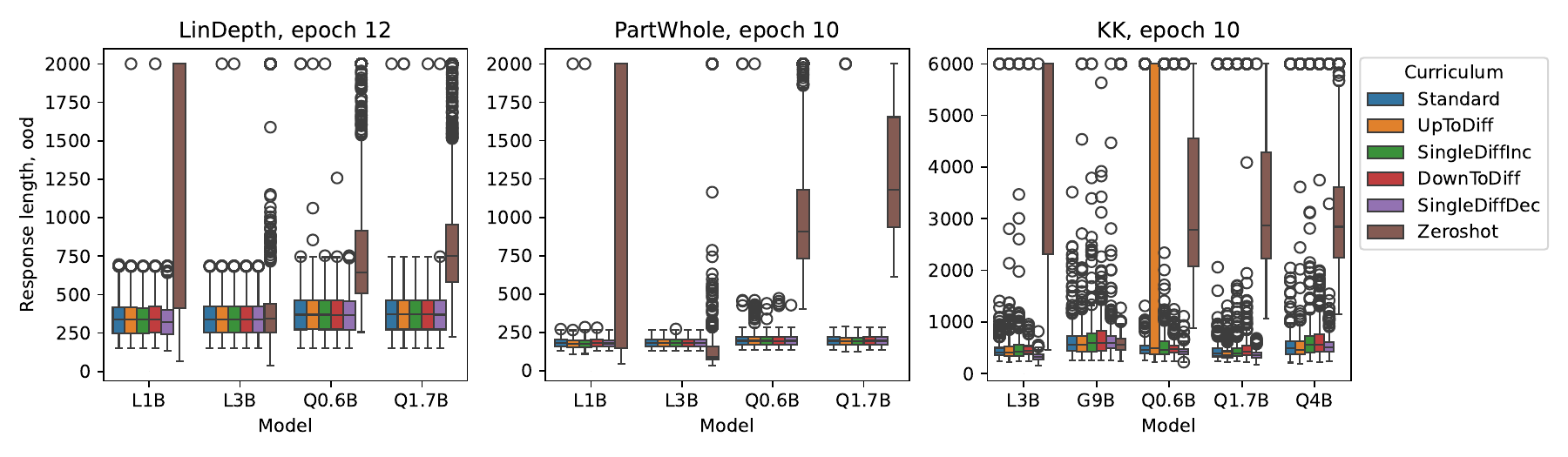}
  \includegraphics[width=\textwidth]{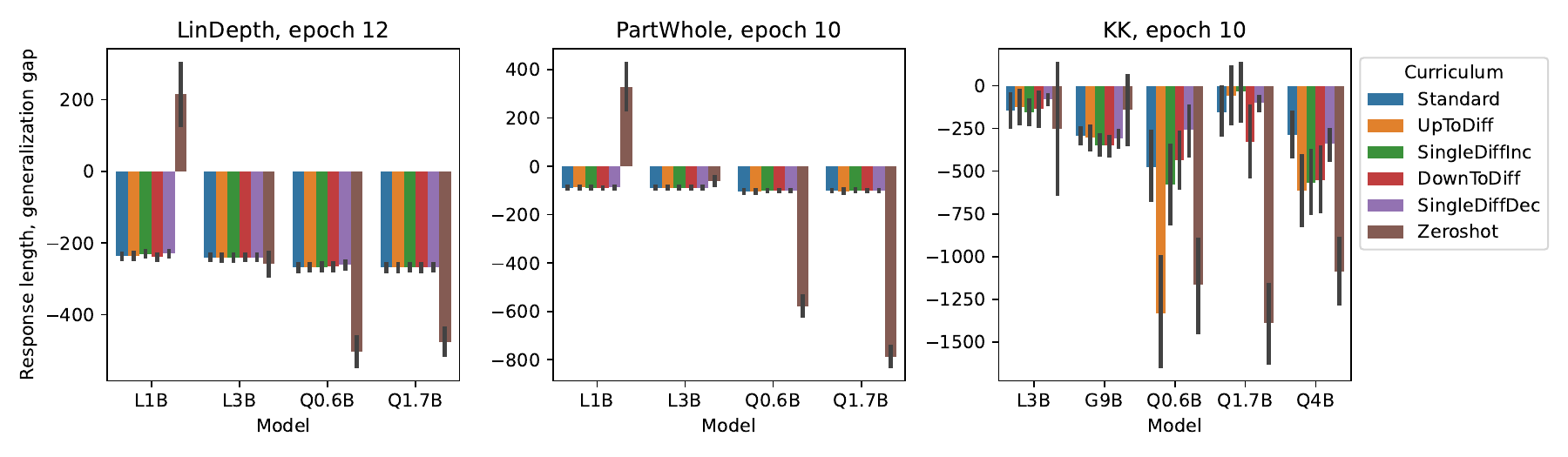}
  \caption{
  \textbf{Response length after SFT post-training.}
  Final-epoch response lengths are shown for in-distribution (\indist) and out-of-distribution (\ood) examples,
  together with the generalization gap (\genGapDef).
  A negative generalization gap indicates that responses are longer on \ood data than on \indist data.
  }
  \label{fig:sft_response_length_metrics}
\end{figure*}

\subsection{Post-Training: Training Evolution}
\subsubsection{GRPO and SFT}
\begin{figure*}[h]
  \centering
  \includegraphics[width=0.45\textwidth]{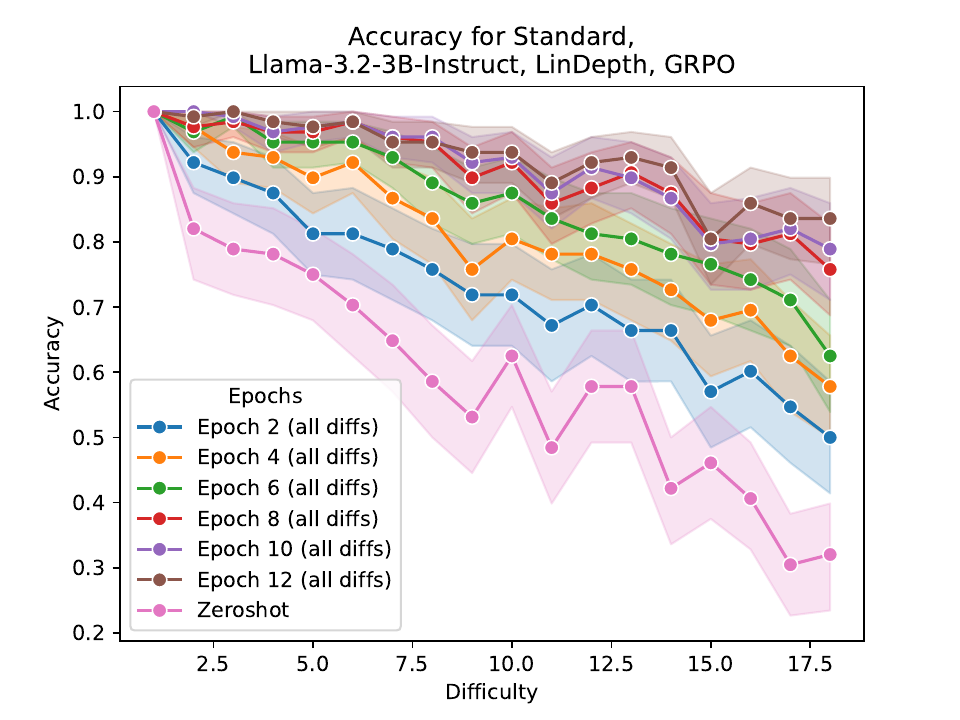}
  \includegraphics[width=0.45\textwidth]{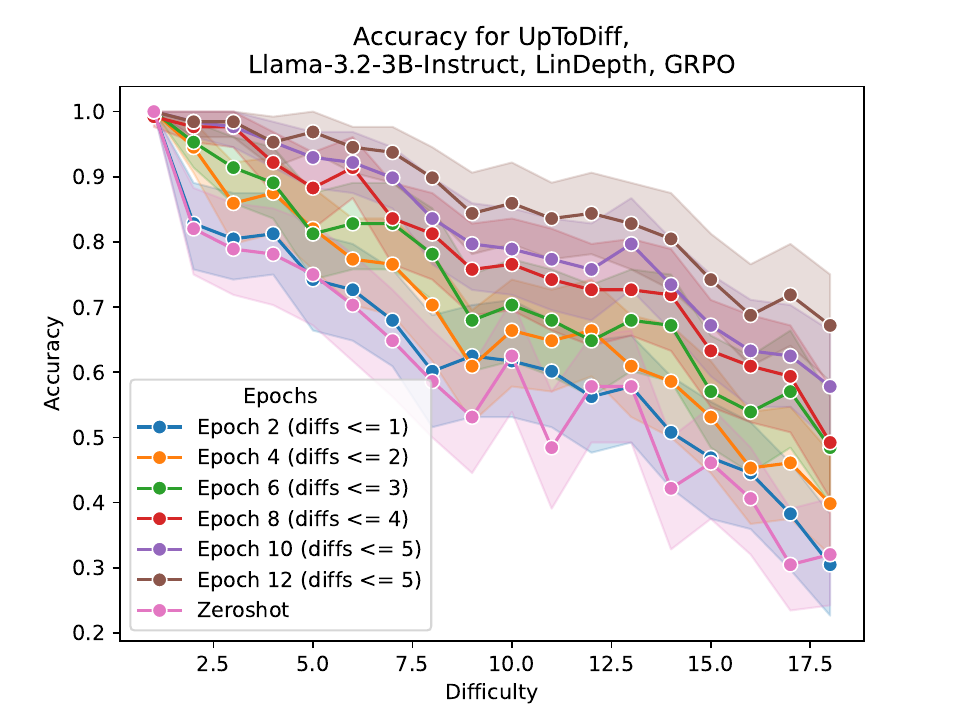}
  \includegraphics[width=0.45\textwidth]{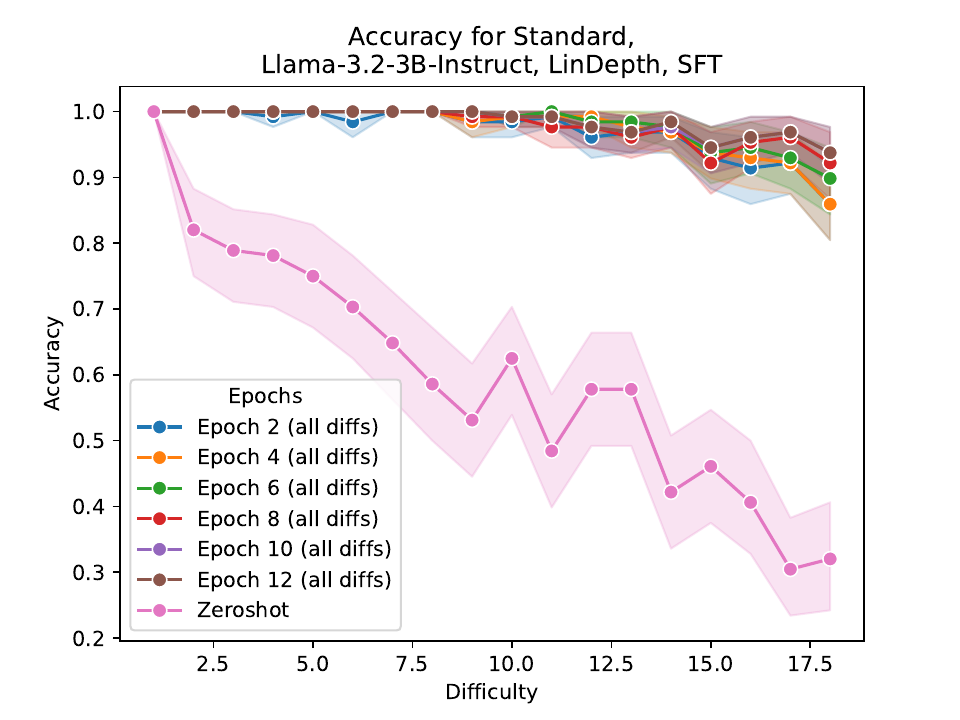}
  \includegraphics[width=0.45\textwidth]{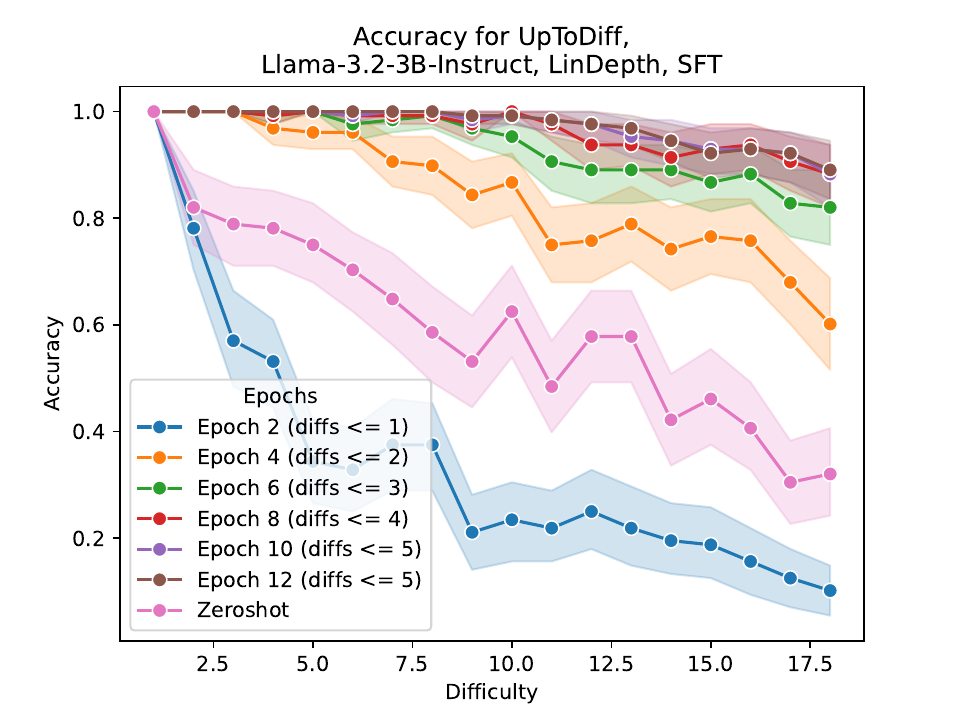}
  \caption{
  \textbf{Training evolution of accuracy on the \lindepth dataset.}
  Accuracy over time across difficulty levels is shown for GRPO (\textbf{top}) and SFT (\textbf{bottom}).
  We illustrate the \noCurr (\textbf{left}) and \upToDiff (\textbf{right}) curricula for model \modelLThree.
  }
  \label{fig:grpo_sft_accuracy_over_time_lindepth}
\end{figure*}

\begin{figure*}[h]
  \centering
  \includegraphics[width=0.45\textwidth]{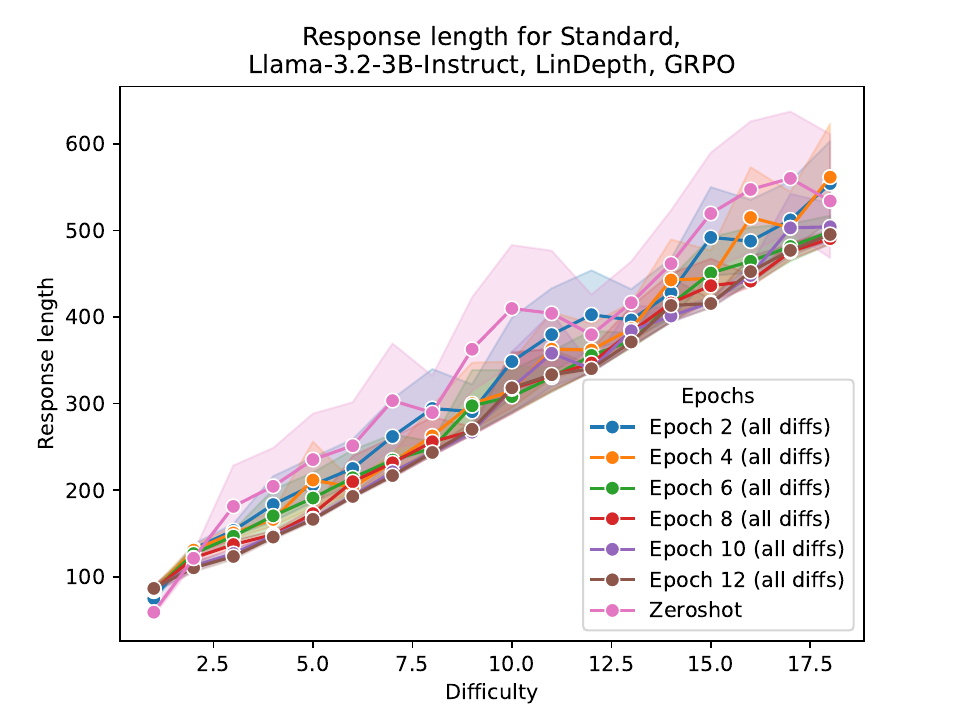}
  \includegraphics[width=0.45\textwidth]{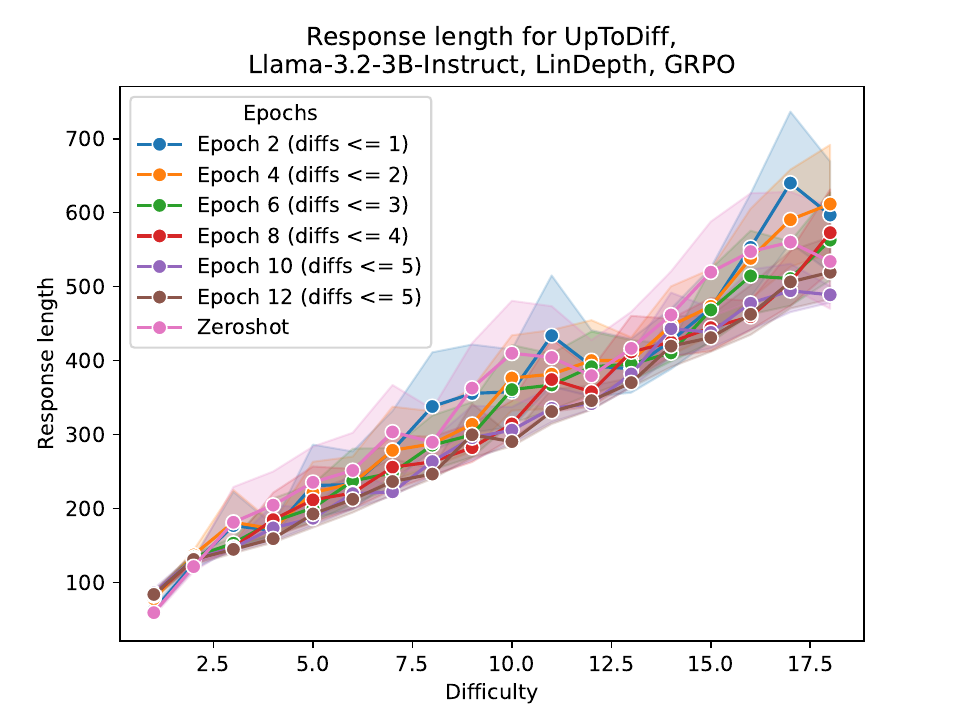}
  \includegraphics[width=0.45\textwidth]{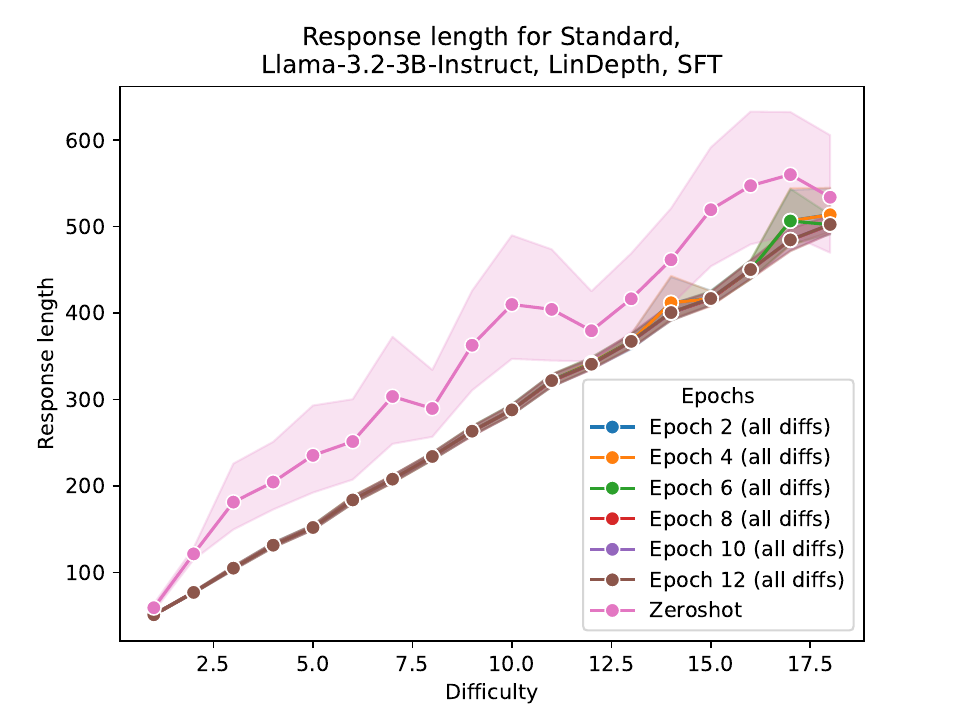}
  \includegraphics[width=0.45\textwidth]{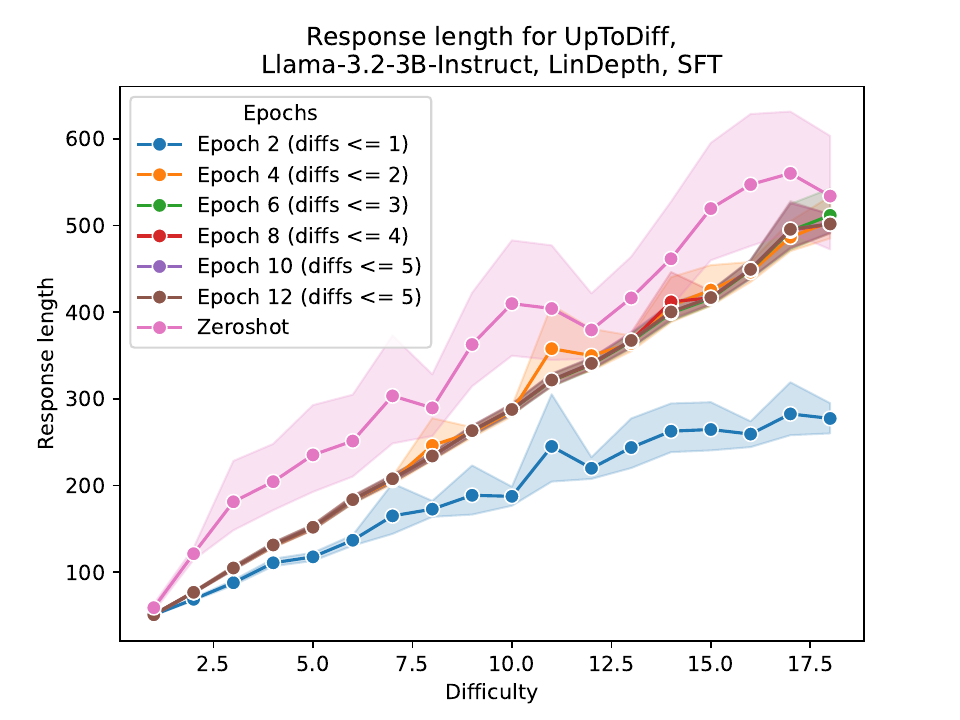}
  \caption{
  \textbf{Training evolution of response length on the \lindepth dataset.}
  Response length over time across difficulty levels is shown for GRPO (\textbf{top}) and SFT (\textbf{bottom}).
  We illustrate the \noCurr (\textbf{left}) and \upToDiff (\textbf{right}) curricula for model \modelLThree.
  }
  \label{fig:grpo_sft_response_length_over_time_lindepth}
\end{figure*}

\begin{figure*}[h]
  \centering
  \includegraphics[width=0.45\textwidth]{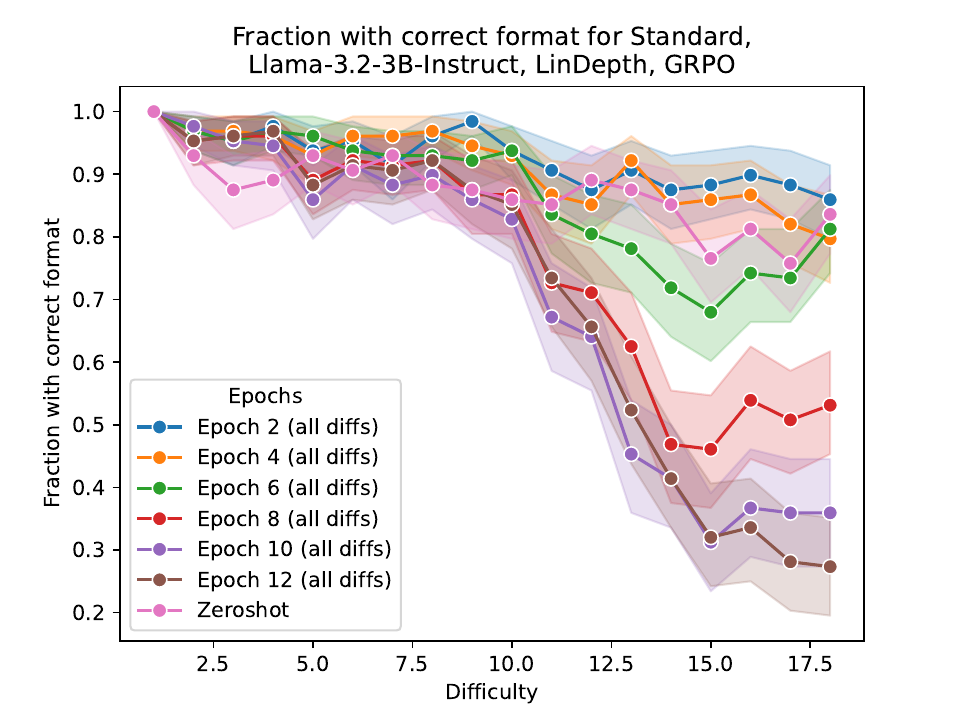}
  \includegraphics[width=0.45\textwidth]{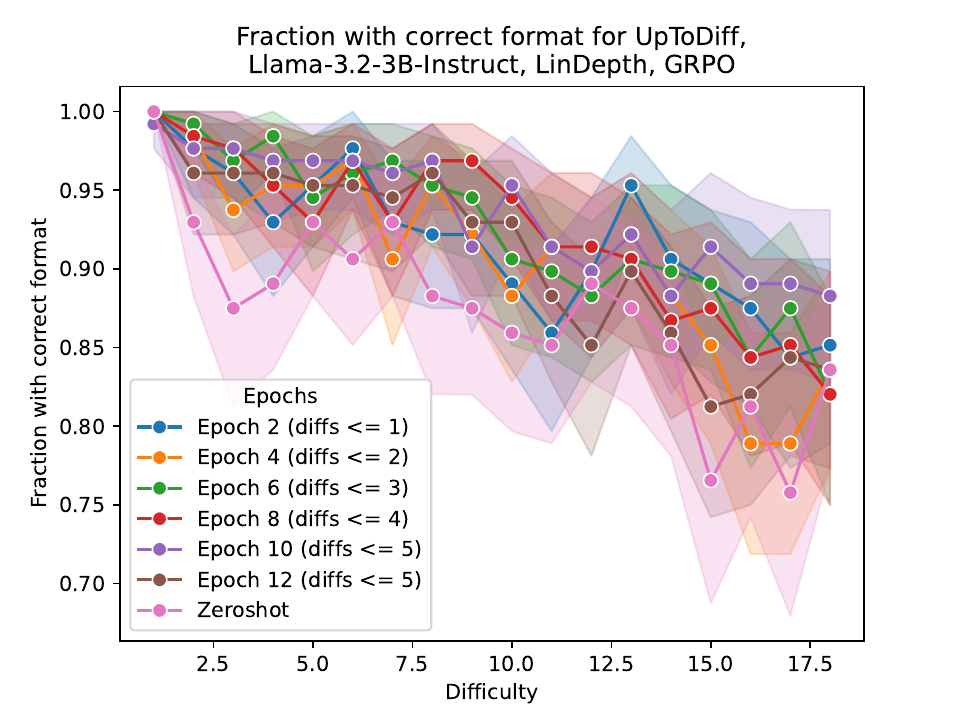}
  \includegraphics[width=0.45\textwidth]{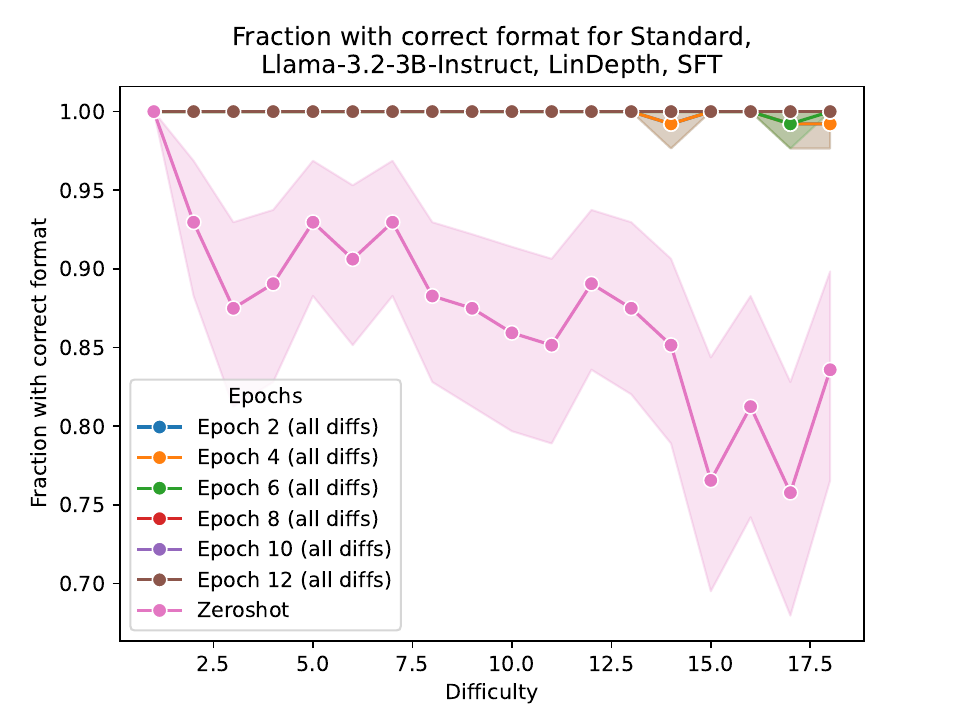}
  \includegraphics[width=0.45\textwidth]{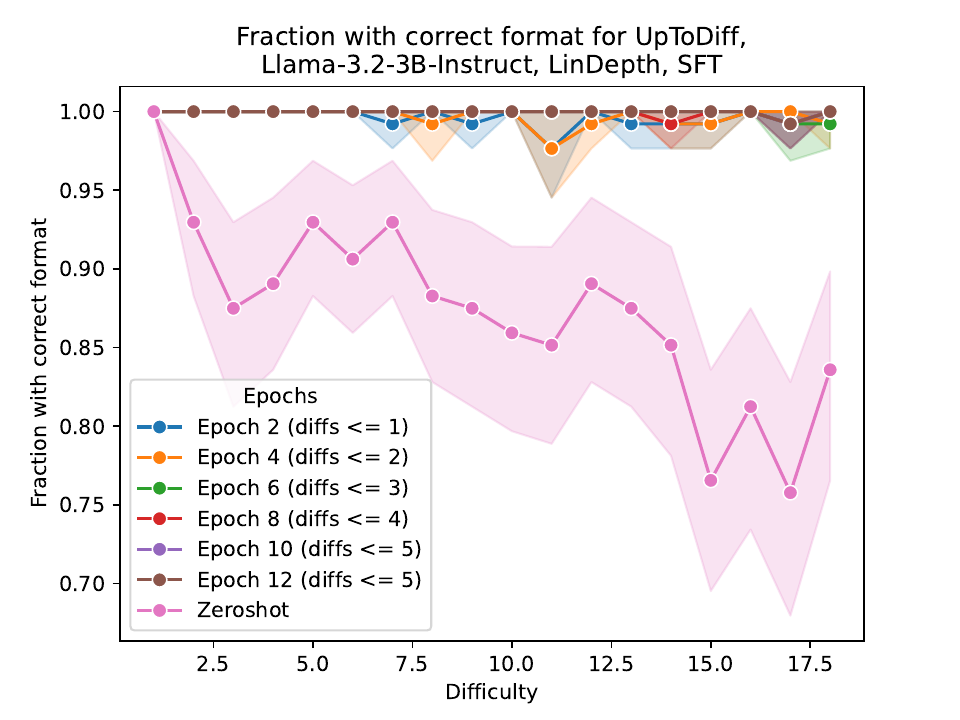}
  \caption{
  \textbf{Training evolution of format adherence on the \lindepth dataset.}
  The fraction of correctly formatted responses over time across difficulty levels is shown for GRPO (\textbf{top})
  and SFT (\textbf{bottom}).
  We illustrate the \noCurr (\textbf{left}) and \upToDiff (\textbf{right}) curricula for model \modelLThree.
  }
  \label{fig:grpo_sft_format_following_over_time_lindepth}
\end{figure*}

\begin{figure*}[h]
  \centering
  \includegraphics[width=0.45\textwidth]{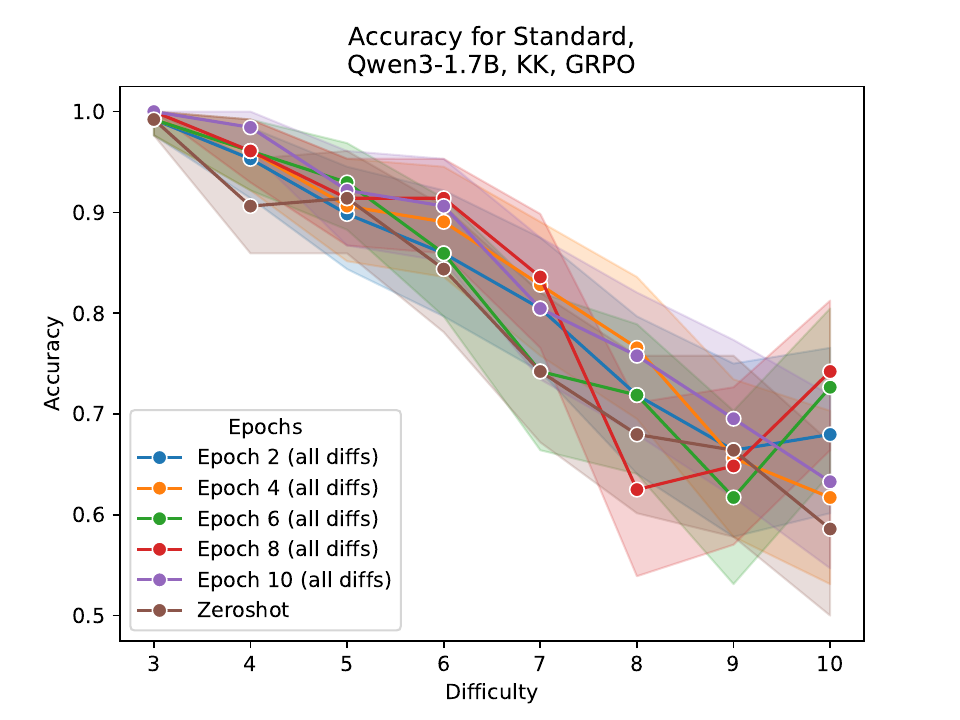}
  \includegraphics[width=0.45\textwidth]{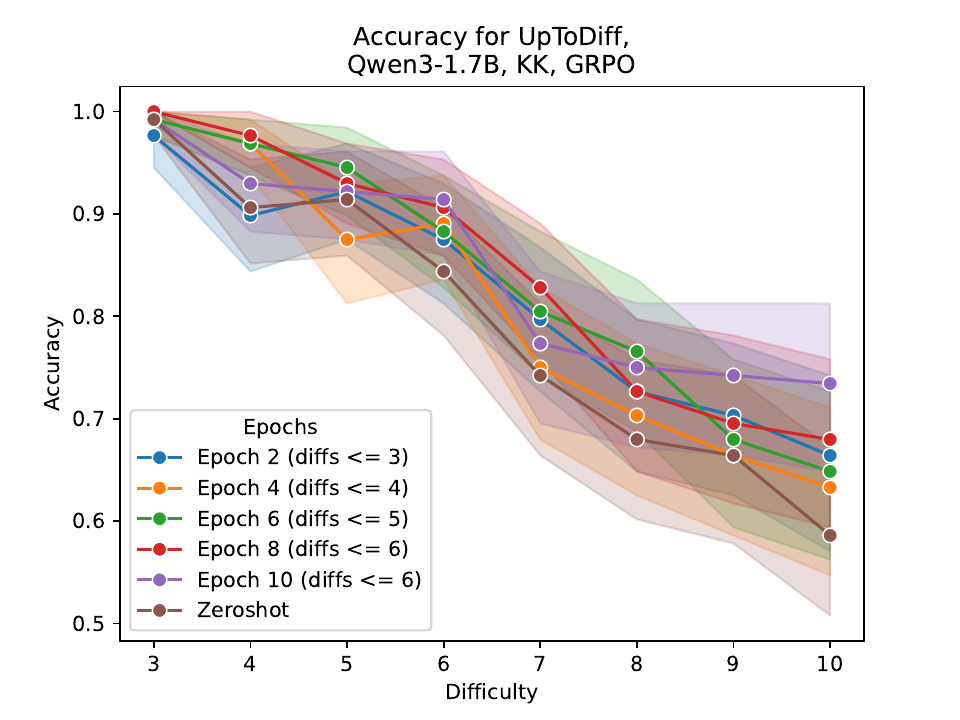}
  \includegraphics[width=0.45\textwidth]{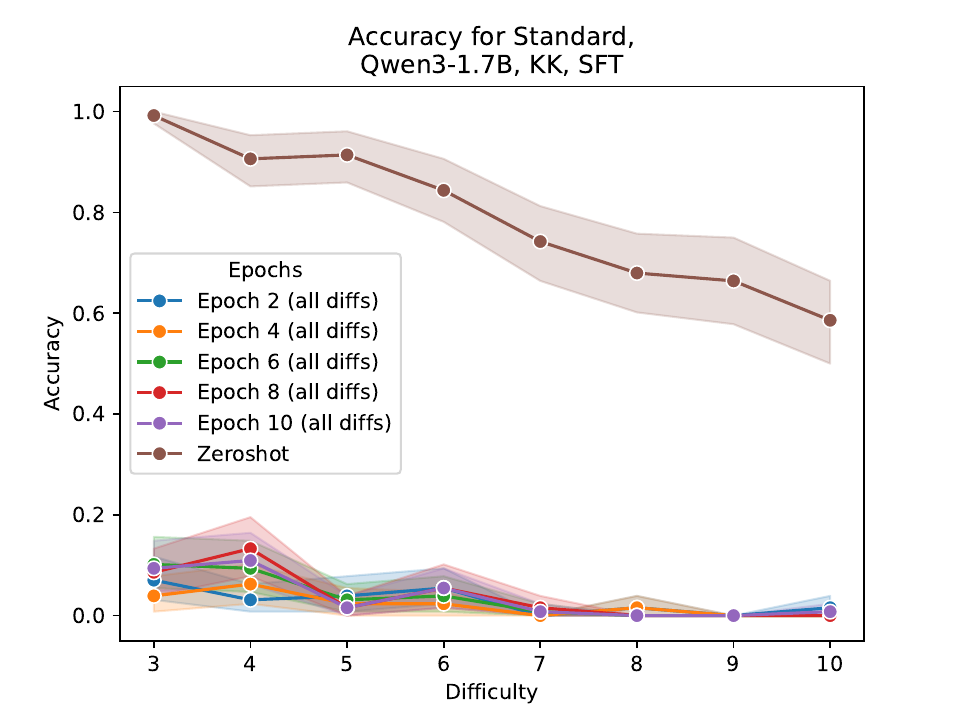}
  \includegraphics[width=0.45\textwidth]{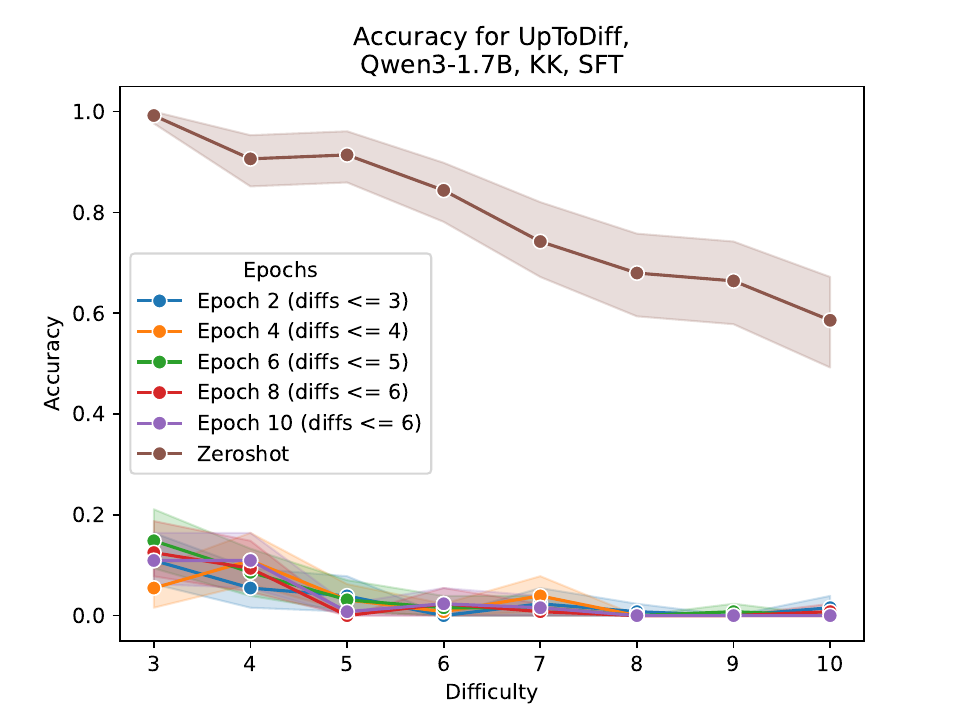}
  \caption{
  \textbf{Training evolution of accuracy on the \kk dataset.}
  Accuracy over time across difficulty levels is shown for GRPO (\textbf{top}) and SFT (\textbf{bottom}).
  We illustrate the \noCurr (\textbf{left}) and \upToDiff (\textbf{right}) curricula for model \modelQwenOneSeven.
  }
  \label{fig:grpo_sft_accuracy_over_time_kk}
\end{figure*}

\begin{figure*}[h]
  \centering
  \includegraphics[width=0.45\textwidth]{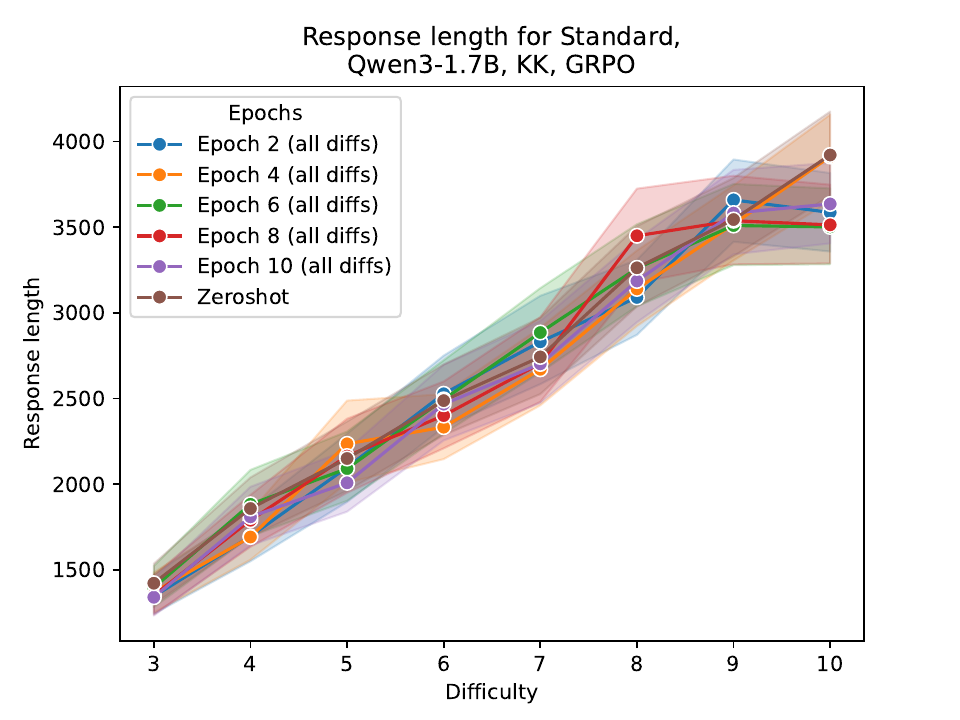}
  \includegraphics[width=0.45\textwidth]{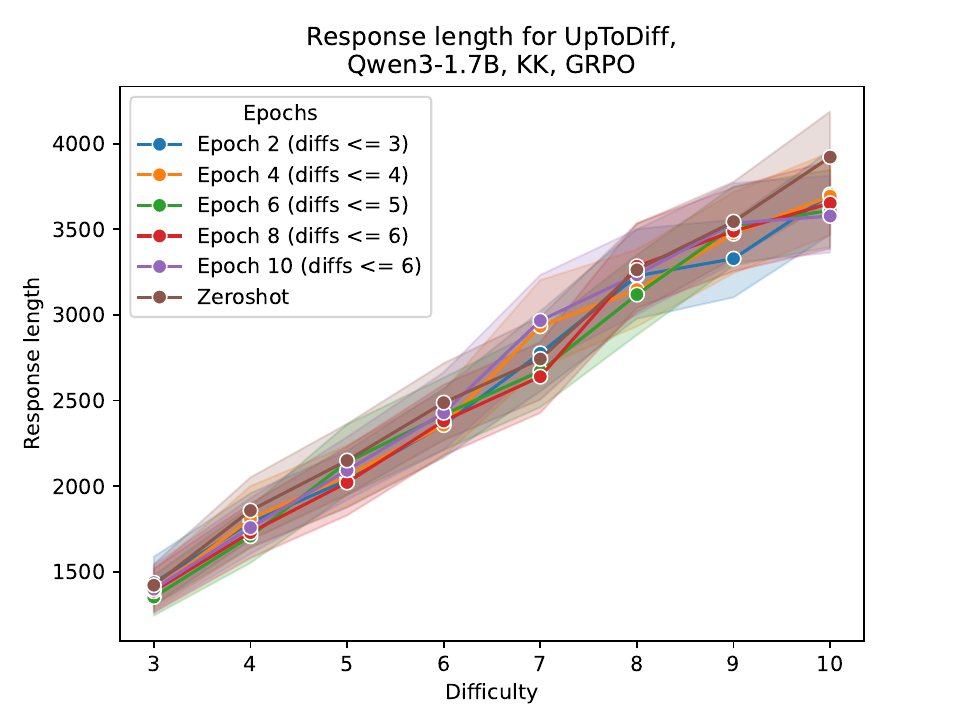}
  \includegraphics[width=0.45\textwidth]{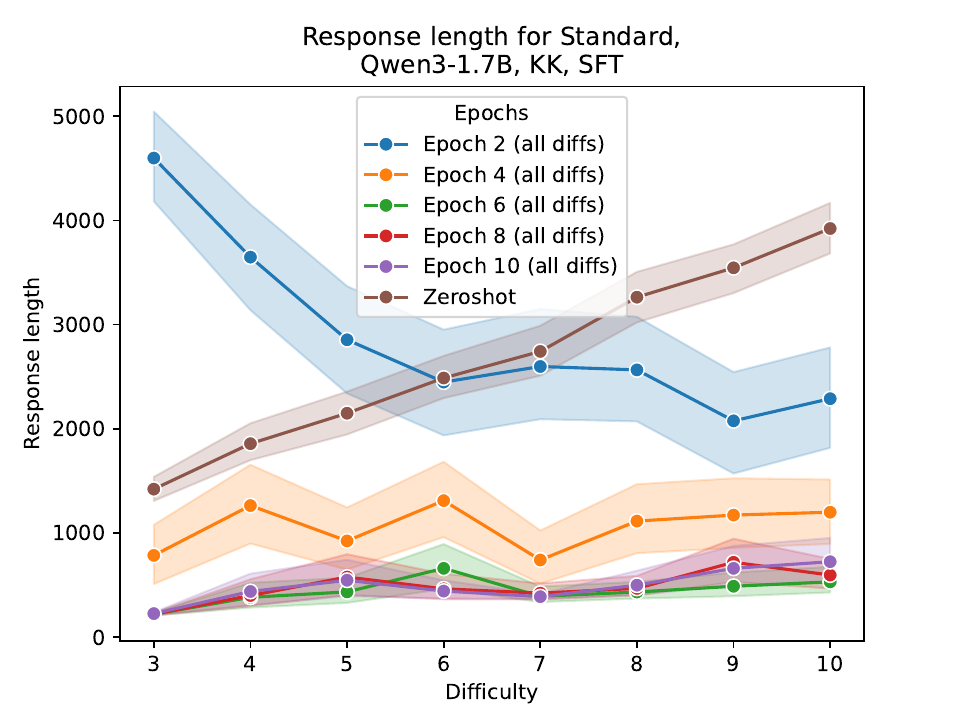}
  \includegraphics[width=0.45\textwidth]{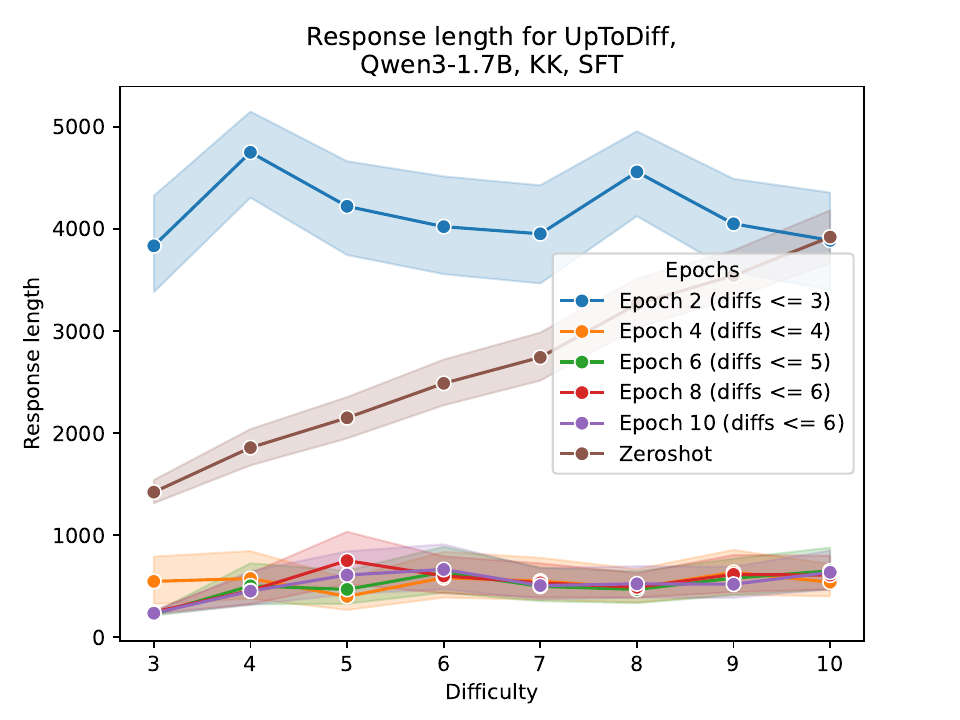}
  \caption{
  \textbf{Training evolution of response length on the \kk dataset.}
  Response length over time across difficulty levels is shown for GRPO (\textbf{top}) and SFT (\textbf{bottom}).
  We illustrate the \noCurr (\textbf{left}) and \upToDiff (\textbf{right}) curricula for model \modelQwenOneSeven.
  }
  \label{fig:grpo_sft_response_length_over_time_kk}
\end{figure*}

\begin{figure*}[h]
  \centering
  \includegraphics[width=0.45\textwidth]{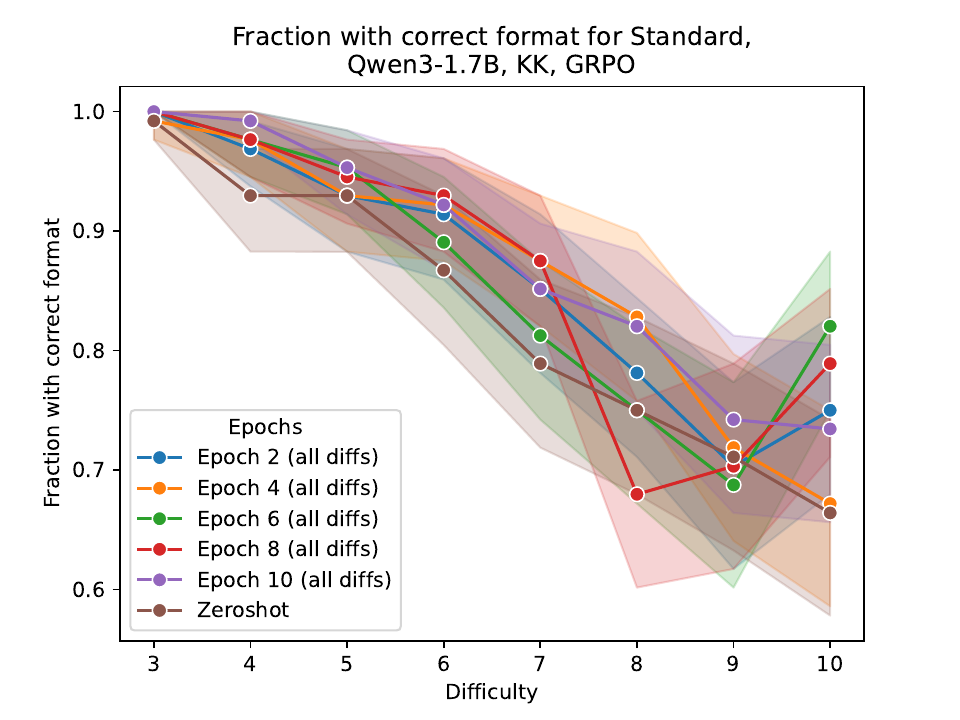}
  \includegraphics[width=0.45\textwidth]{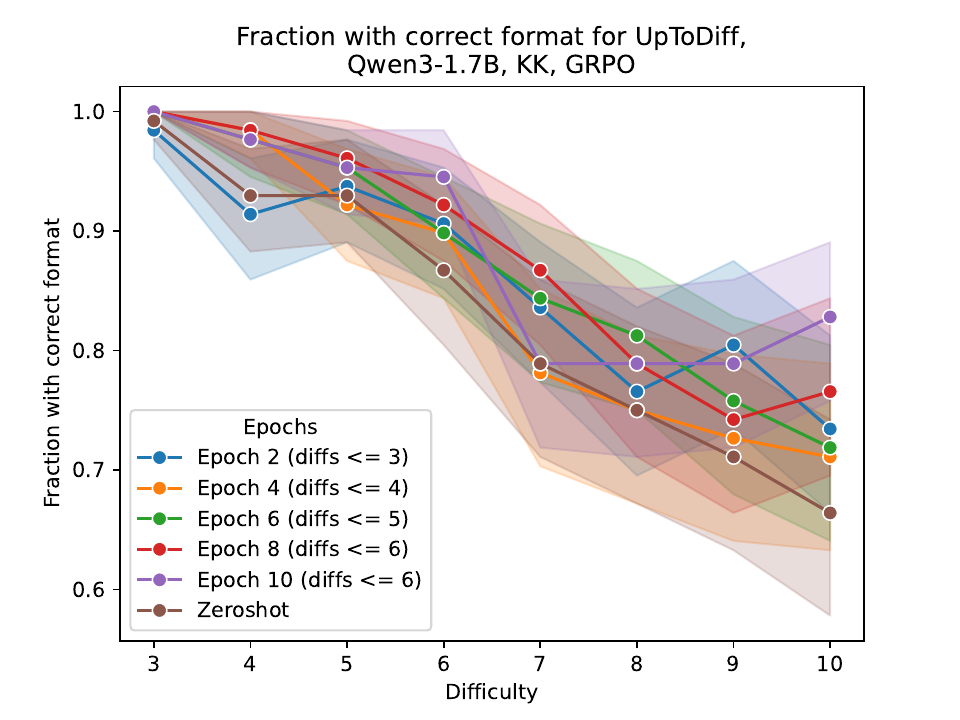}
  \includegraphics[width=0.45\textwidth]{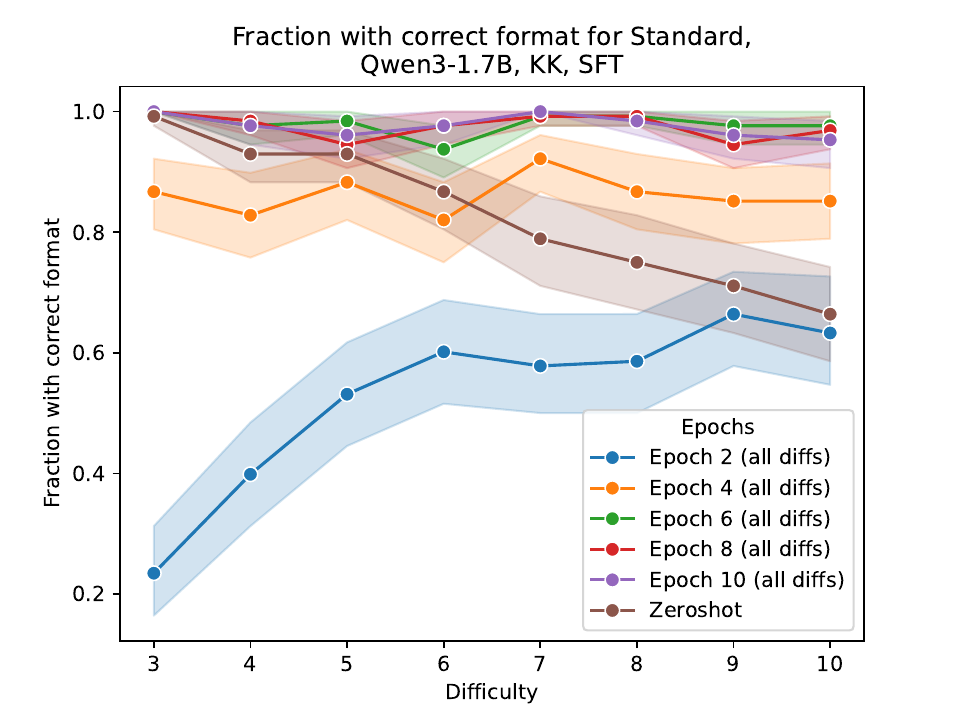}
  \includegraphics[width=0.45\textwidth]{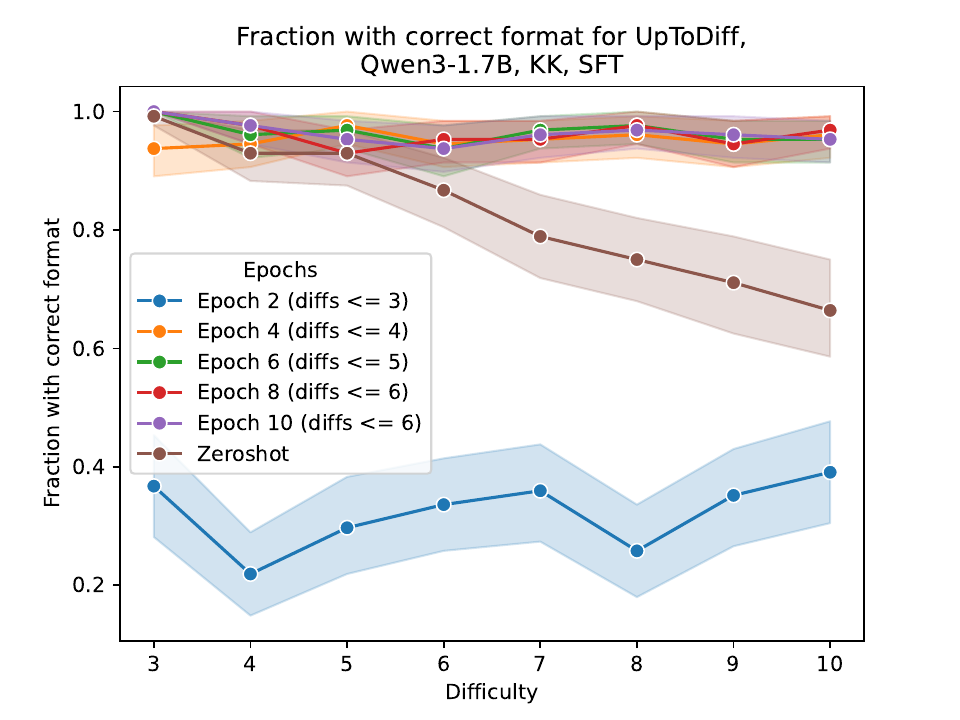}
  \caption{
  \textbf{Training evolution of format adherence on the \kk dataset.}
  The fraction of correctly formatted responses over time across difficulty levels is shown for GRPO (\textbf{top})
  and SFT (\textbf{bottom}).
  We illustrate the \noCurr (\textbf{left}) and \upToDiff (\textbf{right}) curricula for model \modelQwenOneSeven.
  }
  \label{fig:grpo_sft_format_following_over_time_kk}
\end{figure*}

For the \noCurr~and \upToDiff~curricula, \cref{fig:grpo_sft_accuracy_over_time_lindepth,fig:grpo_sft_response_length_over_time_lindepth,fig:grpo_sft_format_following_over_time_lindepth,fig:grpo_sft_accuracy_over_time_kk,fig:grpo_sft_response_length_over_time_kk,fig:grpo_sft_format_following_over_time_kk} compare GRPO against SFT over time in terms of accuracy, response length and fraction of correctly formatted responses. 
For the \lindepth and \kk~dataset, we show the results for the \modelLThree~and \modelQwenOneSeven~models, respectively. Finetuning improves performance significantly on \lindepth, less so on \kk. No consistent difference between the curricula can be observed.

\begin{figure*}[h]
  \centering
  \includegraphics[width=0.45\textwidth]{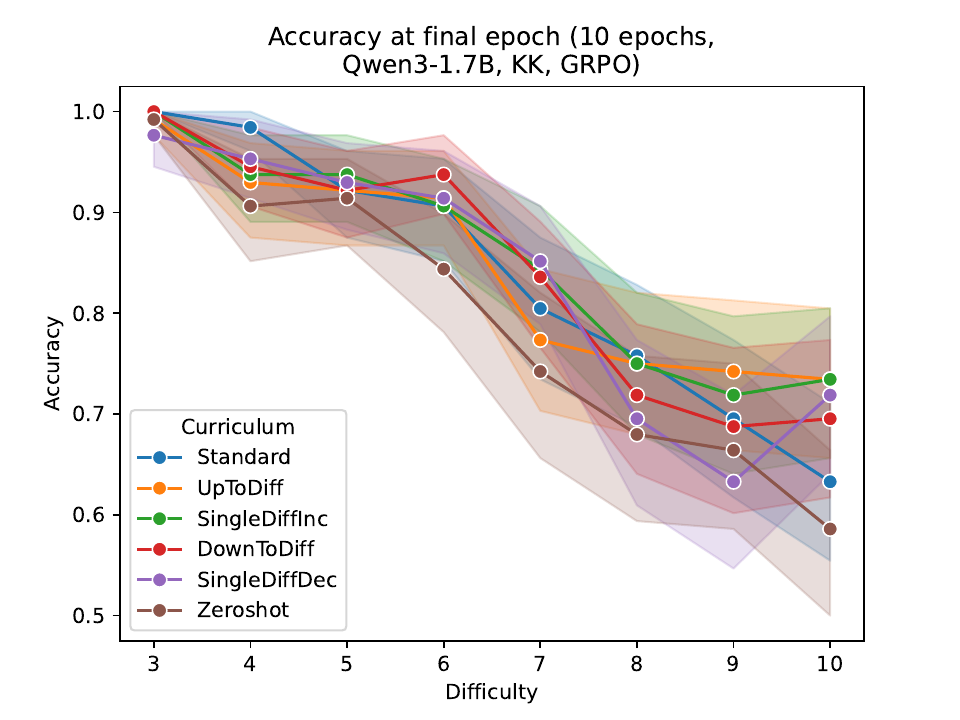}
  \includegraphics[width=0.45\textwidth]{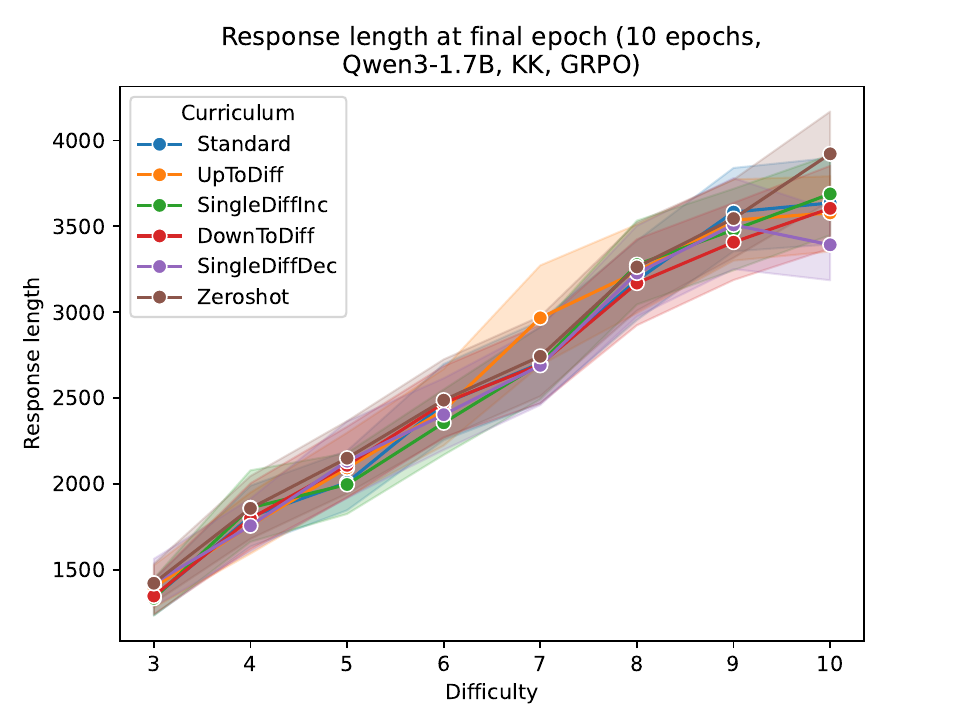}
  \includegraphics[width=0.45\textwidth]{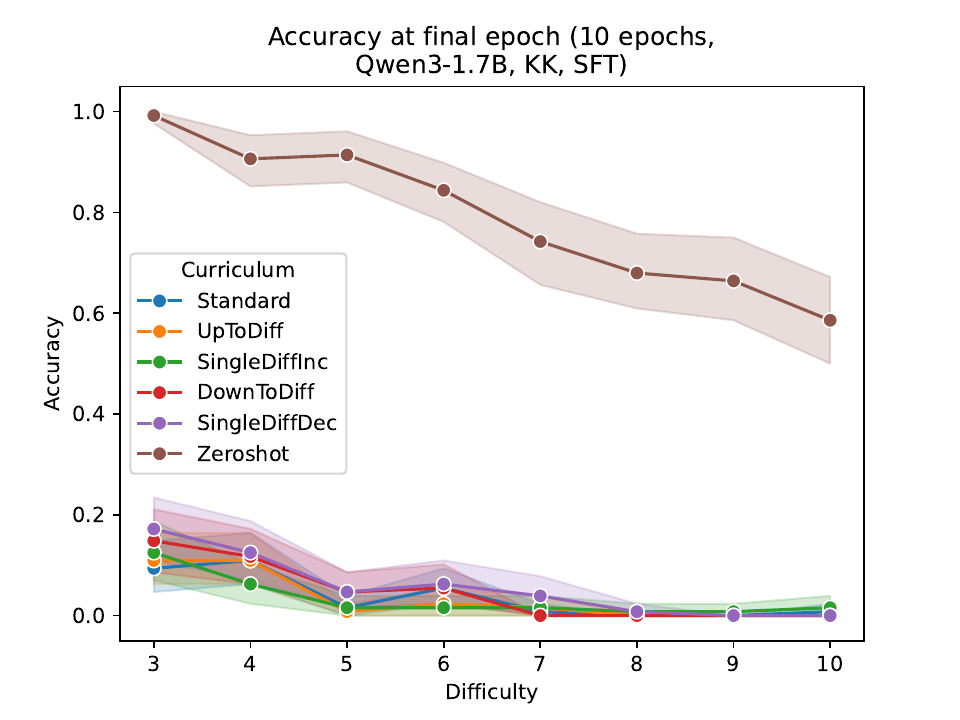}
  \includegraphics[width=0.45\textwidth]{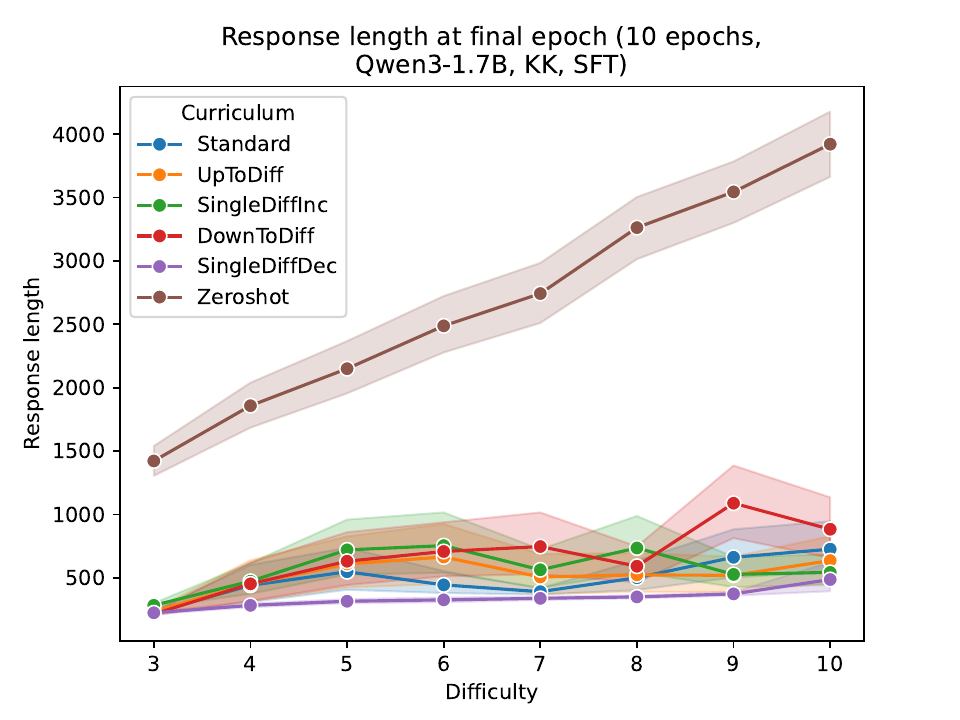}
  \caption{
  \textbf{Accuracy and response length at the final epoch on the \kk dataset.}
  Final-epoch accuracies and response lengths are shown for GRPO (\textbf{top}) and SFT (\textbf{bottom}).
  }
  \label{fig:grpo_sft_response_length_across_curricula}
\end{figure*}

\Cref{fig:grpo_sft_response_length_across_curricula} compares GRPO against SFT across curricula at the final epoch.

\subsubsection{PPO}
\begin{figure*}[h]
  \centering
  \includegraphics[width=0.45\textwidth]{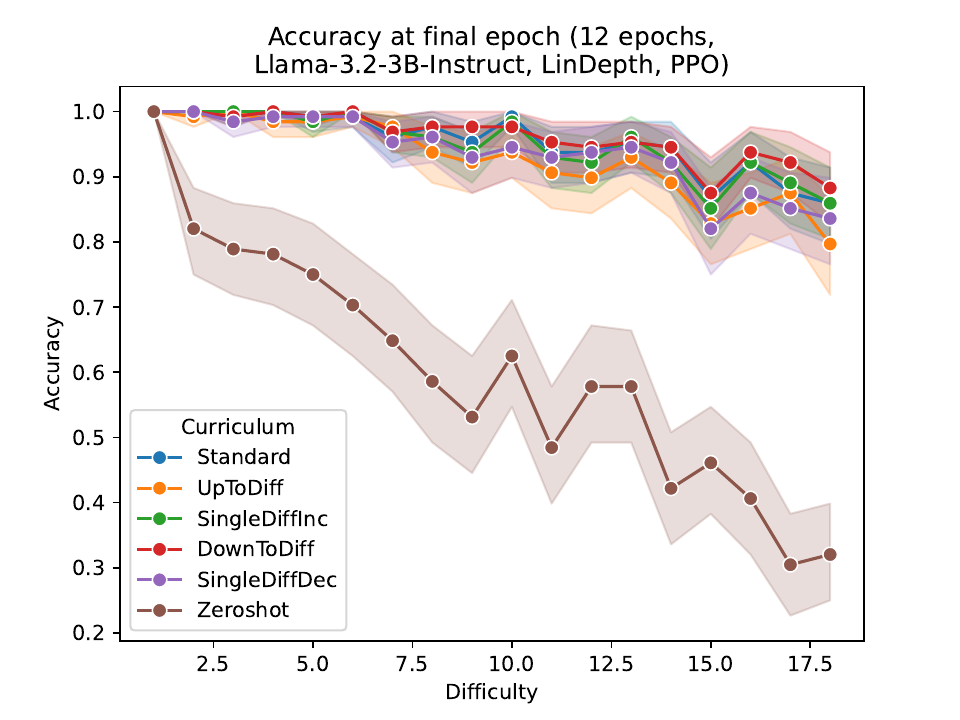}
  \includegraphics[width=0.45\textwidth]{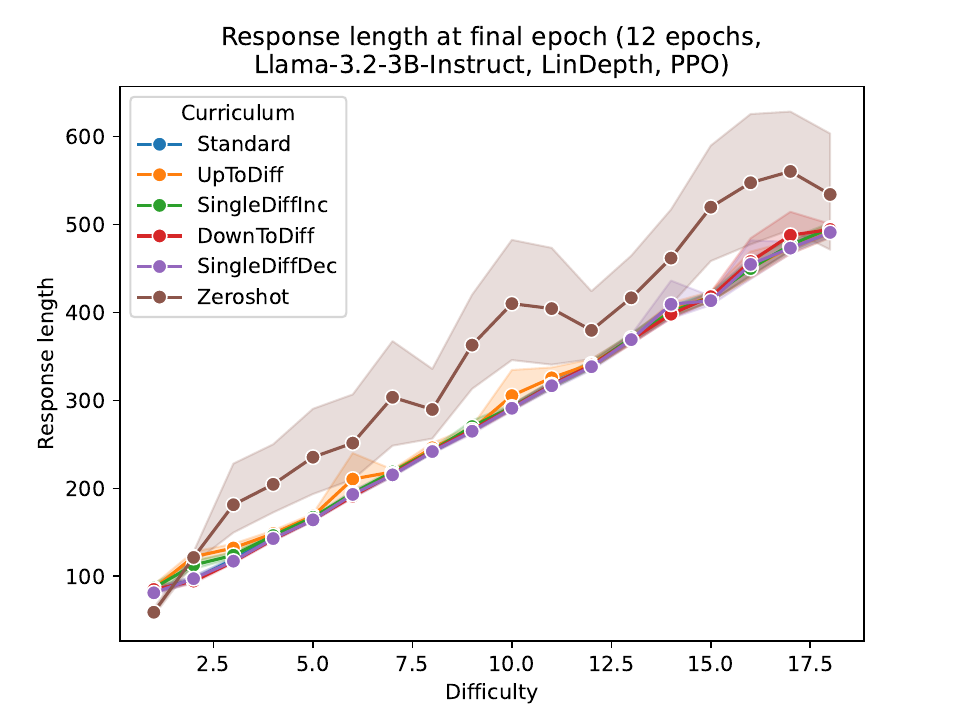}
  \includegraphics[width=0.45\textwidth]{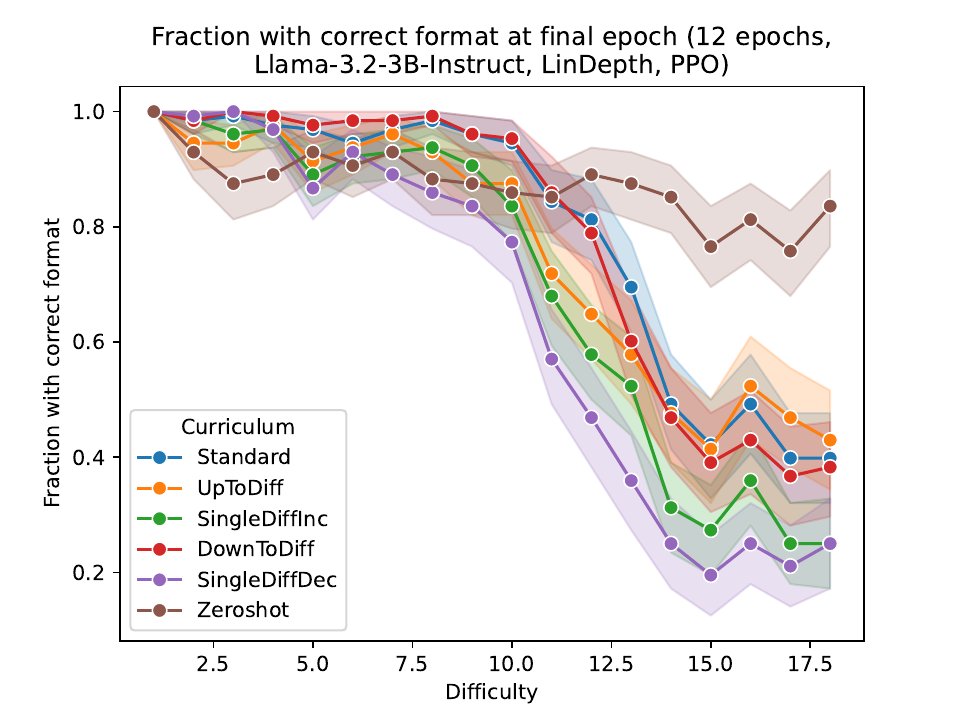}
  \caption{
  \textbf{Post-training performance after PPO on the \lindepth dataset.}
  Final-epoch accuracy, response length, and fraction of correctly formatted responses are shown per difficulty
  for different curricula.
  While format adherence degrades at higher difficulty levels, answer accuracy remains high.
  }
  \label{fig:ppo_metrics_lindepth}
\end{figure*}

We show similar results for PPO in \cref{fig:ppo_metrics_lindepth} at the final epoch for different curricula, per difficulty.

\end{appendices}


\end{document}